%% file: main.tex
\pdfoutput=1
\documentclass[11pt]{article}
\usepackage{enumerate}
\usepackage[OT1]{fontenc}

\usepackage{mystyle}

\usepackage[textsize=tiny]{todonotes}

\newcommand{\RN}[1]{
  \textup{\uppercase\expandafter{\romannumeral#1}}
}

\definecolor{ultramarine}{rgb}{0.,0.1,0.9}
\usepackage[frozencache=true,cachedir=minted-cache]{minted} 

\usepackage{paralist}
\usepackage{fullpage}

\usepackage[utf8]{inputenc}
\usepackage[T1]{fontenc}
\usepackage{url}
\usepackage{booktabs}
\usepackage{amsfonts}
\usepackage{nicefrac}
\usepackage{microtype}
\usepackage{xcolor}
\usepackage{listings}
\usepackage{wrapfig}
\definecolor{darkgreen}{RGB}{1,130,32}
\let\oldthanks\thanks
\renewcommand{\thanks}[1]{\let\footnotemark\relax\oldthanks{#1}}
\usepackage{caption}
\usepackage{minted}
\usepackage{fancyvrb}
\RecustomVerbatimCommand{\VerbatimInput}{VerbatimInput}
{fontsize=\footnotesize,
 breaklines=true,
 breakanywhere=true, 
 breaksymbol=,
 frame=single,  
 framesep=0.7em,
 labelposition=topline,
}
\DeclareMathAlphabet{\mathmybb}{U}{bbold}{m}{n}

\definecolor{darkgreen}{RGB}{1,170,32}
\definecolor{lightblue}{RGB}{1,122,190}

\usepackage{undertilde}
\theoremstyle{plain}

\title{Hindsight Planner: A Closed-Loop Few-Shot Planner for Embodied Instruction Following}

\author{Yuxiao Yang\textsuperscript{1}\qquad 
Shenao Zhang\textsuperscript{2}\qquad
 Zhihan Liu\textsuperscript{2}\qquad 
 Huaxiu Yao\textsuperscript{3}\qquad 
 Zhaoran Wang\textsuperscript{2}\qquad\\
 \small
 \textsuperscript{1}Shanghai Jiao Tong University~~~~~~~~
\textsuperscript{2}Northwestern University~~~~~~~~
\textsuperscript{3}UNC-Chapel Hill
}

\begin{document}

\maketitle

\input{abstract}

\input{intro}
\input{relatework}
\input{motivation}

\input{method}

\input{experiment}

\input{conclusion}

\bibliographystyle{ims}
\bibliography{reference.bib}

\newpage
\appendix
\input{appendix}

\end{document}

%% file: abstract.tex
\begin{abstract}
This work focuses on building a task planner for Embodied Instruction Following (EIF) using Large Language Models (LLMs). Previous works typically train a planner to imitate expert trajectories, treating this as a supervised task. While these methods achieve competitive performance, they often lack sufficient robustness. When a suboptimal action is taken, the planner may encounter an out-of-distribution state, which can lead to task failure. In contrast, we frame the task as a Partially Observable Markov Decision Process (POMDP) and aim to develop a robust planner under a few-shot assumption. Thus, we propose a closed-loop planner with an adaptation module and a novel hindsight method, aiming to use as much information as possible to assist the planner. Our experiments on the ALFRED dataset indicate that our planner achieves competitive performance under a few-shot assumption. For the first time, our few-shot agent's performance approaches and even surpasses that of the full-shot supervised agent.
\end{abstract}

%% file: intro.tex
\section{Introduction}
With the development of AI and robotics, many previous works have combined them to handle Embodied Instruction Following (EIF). Among them, the \textit{Action Learning From Realistic Environments and Directives} (ALFRED) benchmark \citep{ALFRED} is particularly challenging because it requires an agent to learn a long-horizon policy that maps egocentric images and language instructions into a sequence of actions. In each task, the agent will be given a natural instruction (e.g. ``Put a heated mug down on a table'') and an egocentric visual observation at each step. The agent is required to output low-level actions (e.g. MoveAhead, RotateRight, etc.) based on the observation to complete the task. These tasks are usually challenging due to the sparse reward settings. For such a reason, many works have adopted a hierarchical structure to deal with it \citep{LLMplanner,FILM,HLSM,CAPEAN}. The high-level module decomposes the whole task into several sub-goals, the low-level module outputs actions to finish each sub-goal. Previously, sub-goal planners are trained on human-annotated dataset through supervised learning. However, they require large amounts of data and often lack robustness \citep{FILM,HLSM,CAPEAN}.

With recent advancements in Large Language Models (LLMs), many studies have explored using LLMs as sub-goal planners, utilizing their in-context learning abilities \citep{LLMplanner,Socraticplanner,ahn2022icanisay}. Although these methods have achieved competitive performance under the few-shot assumption, a critical limitation is that these approaches all study the problem from a supervised learning perspective. They merely attempt to imitate the ground truth trajectories, which results in a lack of robustness within their agents. EIF benchmarks, on the other hand, require long-horizon planning ability. For example, the task ``Put a warmed apple in the fridge'' requires 12-step planning. Assuming that after applying in-context learning, the distribution of the agent's output actions becomes closer to that of the Oracle, with an accuracy of 0.9, the overall accuracy of the entire planning task decreases to $0.9^{12} = 0.28$. Traditionally, a large amount of data is required to mitigate such an issue \citep{HLSM,CAPEAN}. However, under the few-shot assumption, in-context learning methods rely heavily on the reasoning ability of pretrained LLMs \citep{llm_few_shot_learner,ICL_servey}. The hallucination problem of LLMs \citep{zhang2023sirenssongaiocean} suggests that supervised methods through in-context learning are limited.

To address this issue, we approach the ALFRED task \citep{ALFRED} as a Partially Observable Markov Decision Process (POMDP), where the planner makes decisions based on its current state. Each task begins with a natural language description. At each step, the planner receives an egocentric RGB image and returns a high-level sub-goal. The planner can only receive reward signals (Success or Fail) at the end of the task. There are three major challenges in building a robust planner: (1) The sparse reward settings make it difficult for the planner to learn and make accurate decisions. (2) The planner can only receive an egocentric picture and cannot detect the whole state. (3) Under the few-shot assumption, the planner cannot obtain enough information from trajectories.

For the first problem, we adopt an actor-critic framework \citep{RAFA} which consists of two actors, one critic, and one  generator. At each step, the planner receives a new state and performs a tree search with the actors and generator to plan future trajectories, rather than directly outputting a sub-goal. The critic is then used to select the best rollout and return its initial action. Thus, the planner can optimize the output over the long horizon to address the issue of sparse reward. For the second difficulty, we design an adaptation module instantiated by LLMs. Upon receiving an egocentric image, the adaptation module aims to predict the invisible latent PDDL variables of the task, which could help the planner better understand the environment. For the third challenge, we propose a novel hindsight method. It collects suboptimal trajectories from the agent in the training environment and relabels them to complete the task. This approach provides the planner with additional information. During the deployment phase, we prompt one actor with ground truth samples, while the other actor is prompted with hindsight samples. Thus, the relabeled trajectories can guide the planner in adjusting its policy when incorrect actions are proposed and executed.

\vspace{-0.3cm}
\begin{figure}[H]
\centering
\includegraphics[width=\textwidth]{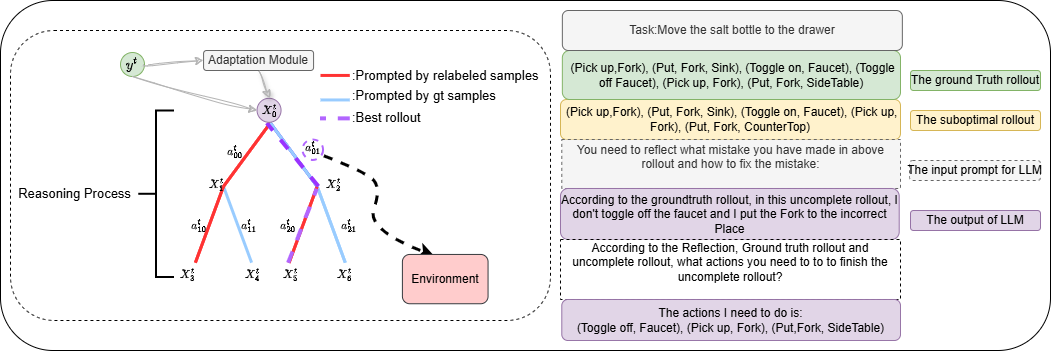}
\setlength{\belowcaptionskip}{-10pt}
\vspace{-0.7cm}
\caption{\small
\textbf{Left}: The illustration of the Hindsight Planner: at each time step $t$, the planner receives a partial observation $y^t$ from the environment. The adaptation module estimates the latent variable and concatenates it with $y^t$ to produce the complete state $x_t$. $\texttt{Actor}_\texttt{hind}$ and $\texttt{Actor}_\texttt{gt}$ are prompted with different samples and make decisions. The \texttt{Critic} is utilized to evaluate the actions. The best rollout $(x_t,a_t^*,x_{t+1}^*,a_{t+1}^*\ldots)$ is selected, and $a_t^*$ is returned. \textbf{Right}: An example of the relabeling process for the $\texttt{Actor}_\texttt{hind}$: after collecting a suboptimal rollout, the LLM is prompted to generate a reflection on the previously taken actions. Following this reflection, the LLM is then prompted to complete the suboptimal rollout.}
\label{fig:demo2}


\end{figure}

In summary, our contributions are threefold:

(1) We study ALFRED \citep{ALFRED} from a POMDP perspective for the first time and propose a closed-loop actor-critic planner to solve it.

(2) We propose a novel hindsight prompting method and demonstrate that our method is theoretically superior to previous approaches.

(3) Experiments on ALFRED \citep{ALFRED} show that our method achieves state-of-the-art performance under few-shot assumptions.
Specifically, the success rates for the ``Test Seen'' and ``Test Unseen'' splits are $25.51$ and $18.77$, respectively, representing a $60\%$ and $39\%$ improvement over the previous baseline.

%% file: relatework.tex
\section{Related Work}
\subsection{Large Language Model (LLM) and In-Context Learning (ICL)}
Large language models (LLMs) have shown incredible reasoning ability \citep{attention,Emergent_Abilities,Llama,gpt4} across a wide range of tasks. A crucial way to enhance this reasoning ability is through in-context learning (ICL) \citep{llm_few_shot_learner,ICL_servey}, which allows LLMs to solve complex tasks with only a few samples. Furthermore, this approach removes the need for fine-tuning, which can be time-consuming and computationally expensive. To utilize the ICL ability better, many studies propose certain frameworks aimed at enhancing the reasoning capabilities of LLMs \citep{TOT,COT,GOT}. Among them, \cite{RAFA} proposes a novel perspective by bridging RL and LLM, which inspires us to study ICL from an RL aspect. \cite{ICL_explanation} interprets ICL as Implicit Bayesian Inference, while \cite{dai2023gptlearnincontextlanguage} believes that ICL is performing implicit Gradient Descent. All of these imply the importance of the content in ICL, an area that remains relatively understudied. To this end, we propose Hindsight Planner as an exploration.
\subsection{Adaptation Module in POMDP}
In a Partially Observable Markov Decision Process (POMDP), planners are presented with observable states, while the latent states are invisible to the planner. Making decisions with incomplete information is challenging; therefore, a component to map the observable state into the latent space is crucial \citep{pomdphindsight}. Adaptation modules have been proven effective in legged robots \citep{kumar2021rmarapidmotoradaptation,5655cf7f69e149148a2ea1c5a664b4ba,peng2020learningagileroboticlocomotion}. These modules aim to bridge the gap between the simulator and the real world. They are often trained to predict crucial information that a robot can sense in the simulator but not through its sensors in the actual world, such as surface friction or payload of the robot. The base policy then makes decisions based on the observed information and the invisible latent information predicted by adaptation modules. Inspired by this, we propose an adaptation model that maps the visible object list to the latent, invisible Planning Domain Definition Language (PDDL) \citep{PDDL} of ALFRED \citep{ALFRED}.


Previous work such as \cite{FILM}, trains a BERT \citep{BERT} to predict the PDDL arguments and decompose high-level instructions into templated sub-goals. However, our approach differs from these in two aspects: (1) Previous works predict the arguments at the beginning of a task, which is equivalent to predicting the latent variables based on the initial observed state. In contrast, our method predicts the latent arguments at each time before reasoning, allowing predictions to be adjusted through exploration, which makes our planner more robust. (2) We do not apply the templated approach directly. The adaptation module is used to reveal the latent information for the planner and assist the planner in making better decisions. Experiments show that our method achieves competitive performance even without the assistance of the adaptation model, as demonstrated in \Cref{tab:3}.
\subsection{Hindsight in LLMs}
Hindsight algorithms \citep{andrychowicz2018hindsightexperiencereplay,li2020generalizedhindsightreinforcementlearning,pong2020temporaldifferencemodelsmodelfree} are widely adopted in the reinforcement learning (RL) area. Generally, the hindsight method aims to reveal future information after collecting a trajectory and relabel the trajectory to make it more informative during training process \citep{furuta2022generalizeddecisiontransformeroffline,andrychowicz2018hindsightexperiencereplay}. \cite{furuta2022generalizeddecisiontransformeroffline} applies the hindsight method in training a Transformer model and achieves competitive performance on several baselines. However, training a model from scratch usually requires a large amount of data. In contrast, in-context learning, leveraging the reasoning ability of LLMs, allows an agent to complete complex tasks with only a few samples. \cite{dai2023gptlearnincontextlanguage} has shown that ICL executes an implicit parameter update. As a result, we utilize ICL in our proposed method. Intuitively, we hope hindsight prompts can provide guidance when an out-of-distribution state is encountered.  For example, ``Wash a pan and put it away'' requires the agent to wash a \textit{Pan} and put it on the \textit{DiningTable}. The trajectory from a planner could be:
 \{(PickupObject, Pan), (PutObject, Sink),  (ToggleObjectOn, Faucet), (PickupObject, Pan), (PutObject, CoffeeMachine)\}.
 Note that in this example, the agent fails to place the pan in the correct location, does not turn off the faucet, and thus the trajectory from the planner is suboptimal. Our hindsight method proposes a novel relabeling process that appends actions to the suboptimal trajectory, aiming to complete the task. In the above example, the corrected trajectory should be: \{(PickupObject, Pan), (PutObject, Sink),  (ToggleObjectOn, Faucet), (PickupObject, Pan), (PutObject, CoffeeMachine), (ToggleObjectOff, Faucet), (PickupObject,Pan), (PutObject, DiningTable)\}. This approach enables us to guide the planner in addressing unknown states resulting from incorrect actions. Consequently, during the deployment phase, when the planner encounters a similar state, it can learn from suboptimal trajectories and subsequently take correct actions to correct previous mistakes.

We also analyze our method in comparison to previous hindsight methods \citep{andrychowicz2018hindsightexperiencereplay,Ghosh2019LearningTR} following the framework proposed by \cite{furuta2022generalizeddecisiontransformeroffline}. We demonstrate that while previous methods are effective, they alter  the distribution of a crucial variable (the information statistic) in multi-task RL problems. In contrast, our method optimize the same objective while maintaining the distribution. The detailed discussion can be found in \Cref{hindsight}.


%% file: motivation.tex
\section{Preliminaries}
{
 \label{sec:motivation}
\subsection{Definition in POMDP}
In a POMDP~$\mathcal{M}$,  consider an action space $\mathcal{A}$, latent state space $\mathcal{X}$, observation space $\mathcal{Y}$, transition probability function $p(x'|x,a)$, emission function $o(y|x)$, reward function $r(x,a)$ and discount factor $\gamma\in [0,1)$. The trajectory $\tau$ is defined as $\tau = \{x_0,y_0,a_0,x_1,y_1,a_1,\ldots\}$ and the initial state $x_0$ is generated through $x_0\sim \rho_0(\cdot)$.
The policy $\pi_{\theta}(\cdot|y)$ aims to map the observation space into the action space, with $\theta$ denoting its parameters. The goal of RL is to train a policy such that
\begin{align}
    \pi_{\theta} = \arg\max_{\pi}\ \mathbb{E}_{\tau \sim P(\cdot|\pi)}[R(\tau)],
    \label{eq:rl1}
\end{align}
where $P(\tau|\pi)=\rho_0(x_0)\prod_{t=0}^{T}p(x_{t+1}|x_t,a_t)o(y_t|x_t)\pi(a_t|y_t)$, $R(\tau) = \sum_{t=0}^{T} \gamma^t r(x_t,a_t).$

Given a parameterized reward function $r_z(x,a)$, where $z \in \mathcal{Z}$ is a variable indicating the goal for the agent, the conditional policy $\pi(\cdot|y,z)$ aims to accomplish different goals based on its observations. The goal in \Cref{eq:rl1} becomes
\begin{align}
\label{eq:rl2}
    \pi_{\theta} = \arg\max_{\pi}\ \mathbb{E}_{\tau \sim P(\cdot|\pi, z), z\sim p(z)}[R_z(\tau)],
\end{align}
where $R_z(\tau)=\sum_{t=0}^{\infty}\gamma^t r_z(x_t,a_t).$
\Cref{eq:rl2} can be considered as the multi-task RL objective to optimize, which is the core of EIF.
\subsection{Information Matching}
Following \cite{furuta2022generalizeddecisiontransformeroffline}, 
we define the information matching (IM) problem as training a policy $\pi_{\theta}$ that satisfies
\begin{align}
\pi_{\theta} =\arg\min_{\pi} \mathbb{E}_{ \tau\sim P(\cdot|\pi,z),z\sim p(z)}\left[ \text{KL}(I(\tau), z) \right] \label{eq:im},
\end{align}
where $I(\tau)$ is \textit{information statistic} that can be any function that captures the desired information from a trajectory $\tau_t= \{x_0,y_0, a_0, x_1,y_{1}, a_{1}, \dots, x_t,y_{t}\}$ and  $\text{KL}$ is the Kullback-Leibler divergence. This optimization objective has achieved competitive results in previous studies \citep{SMM,hazan2019provablyefficientmaximumentropy}.
\cite{furuta2022generalizeddecisiontransformeroffline} demonstrates that previous hindsight methods \citep{andrychowicz2018hindsightexperiencereplay,eysenbach2020rewritinghistoryinverserl,guo2021hindsightvaluefunctionvariance} utilize various \textit{information statistics} and minimize the divergence $D=0$ by setting  $\hat{z} = I(\tau)$. This allows trajectories to be better used to train a policy $\pi(\cdot| x, z)$. 
For instance, in HER \citep{andrychowicz2018hindsightexperiencereplay}, an MDP trajectory $\tau_{t}^{s} = \{s_0, a_0, s_1, \ldots, s_t\}$ is collected. The information statistic is set as the final state of the agent, where $I(\tau_t^s) = s_t$, and the relabeling process in HER is equivalent to setting $\hat{z} = I(\tau_t^s)$.
}

%% file: method.tex
\section{Hindsight Planner}
\subsection{Overview}

The Hindsight Planner outputs a sub-goal based on the observed objects and natural language instructions. During the collection phase, suboptimal trajectories are collected, and we apply our
\begin{algorithm}[H]\small
	\caption {Hindsight Planner}
 \label{alg:llm}
	\begin{algorithmic}[1]

    \STATE \textbf{Input:} An LLM-planner $\texttt{LLM-PL}$, an adaptation module \texttt{Adapter} and the task instruction  $I$. \STATE \textbf{Set:} Observed Objects $O\leftarrow \emptyset$, the sub-goal history $G\leftarrow\emptyset$, the current sub-goal  $S\leftarrow\emptyset$, the time step $t\leftarrow 0$ and the sub-goal index $k\leftarrow 0$.
    \STATE Get sample pool $\cD$ and initialize  \texttt{Actor}$_{\theta}$,  \texttt{Critic}, \texttt{Adapter} from $\cD$, for any $~\theta\in \{\texttt{gt,hind}\}$ (e.g. \Cref{alg:collection} in \Cref{more-alg}). \hfill (Hindsight process)
 
    \WHILE{Not \textit{Finished}}
        \STATE Get PDDL arguments $P$ $\leftarrow$ $\texttt{Adapter}(I,O)$.
        \STATE Plan and get sub-goal $S_k\leftarrow$  \texttt{LLM-PL}($\texttt{Actor}_\texttt{gt}, \texttt{Actor}_\texttt{hind},\texttt{Critic},P,I,O,G$)(e.g. \Cref{alg: example} in \Cref{more-alg}).
        \STATE Set $S_k$ as sub-goal for \texttt{Low-PL}.
        \WHILE{ $S_k$ not \textit{Finished} and not \textit{Failed} }
            \STATE Invoke \texttt{Low-PL} to plan and execute $a_t$ and update $O$.
            \STATE Set $t\leftarrow t+1$
            
            \IF{$S_k$ \textit{Finished}}
                \STATE Append $S_k$ to $G$.
                \STATE Set $k\leftarrow k+1$.
              
            \ENDIF
        \ENDWHILE

    \ENDWHILE
	\end{algorithmic}
\end{algorithm}
\vspace{-0.5cm}
hindsight method to generate $\cD_{hind}$. The complete dataset $\cD = \cD_\text{hind} \cup \cD_\text{gt}$, where $\cD_\text{gt}$ is constructed from training data. In the deployment phase, we initiate hindsight actor $\texttt{Actor}_\texttt{hind}$, ground truth actor $\texttt{Actor}_\texttt{gt}$, and $\texttt{Critic}$ from $\cD$.

\begin{figure}[H]
\centering
\includegraphics[width=0.8\textwidth]{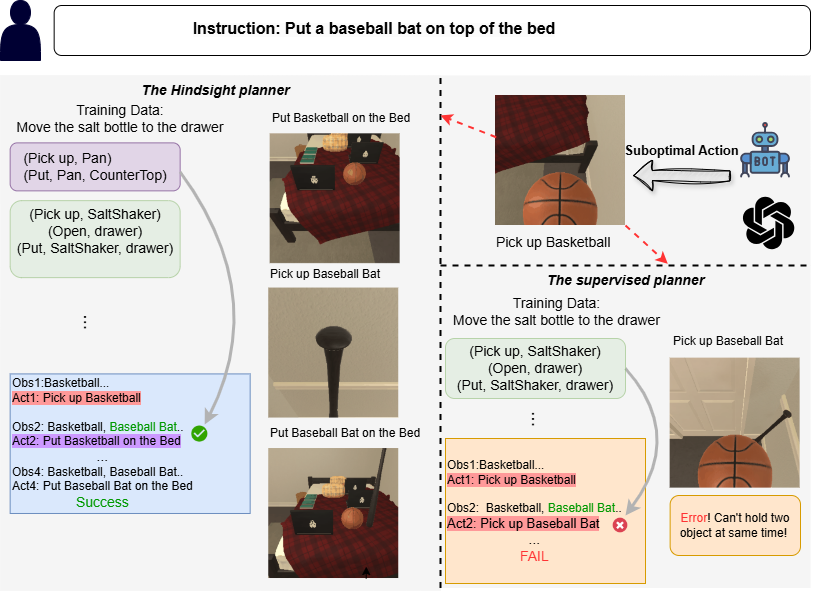}  
\setlength{\belowcaptionskip}{-10pt}
\caption{\small A comparison of Hindsight Planner and previous supervised methods when taking a suboptimal action. The agent initially picks up the incorrect object (``Basketball''). In the supervised method, the planner fails to handle this situation, which leads to task failure. In contrast, the Hindsight Planner can adjust after the incorrect action and successfully complete the task.\vspace{-0.1cm}}
\label{fig:demo1}
\end{figure}

\begin{figure}[H]
\centering
\includegraphics[width=\textwidth]{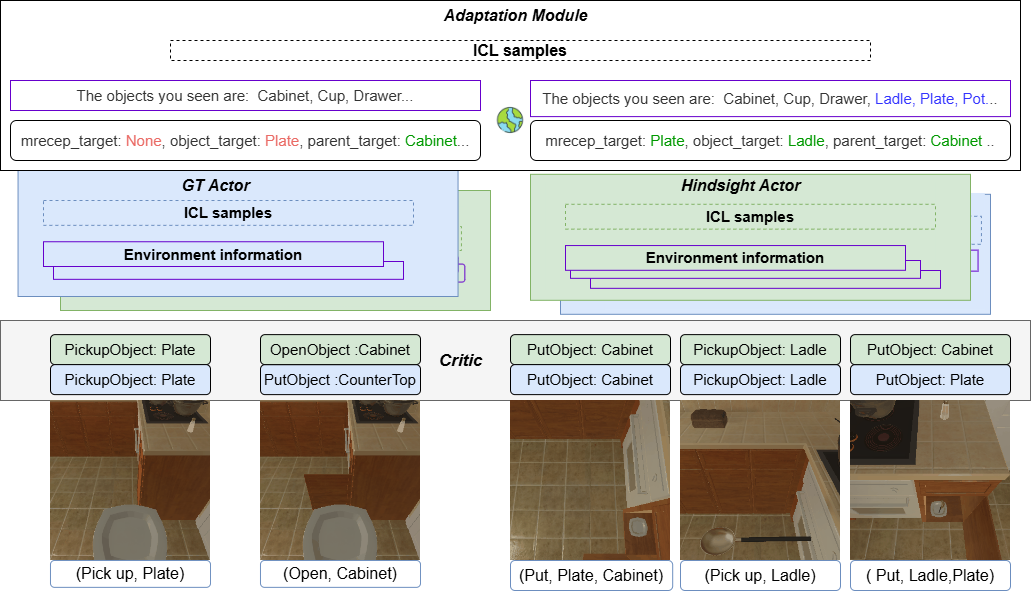}  
\caption{\small The entire process of the Hindsight Planner is as follows: At the start of the task, which is to ``Place a plate with a ladle on it in a cabinet,'' the \texttt{Adapter} mistakenly identifies the task as picking up a plate and placing it into a cabinet. $\texttt{Actor}_\texttt{hind}$ and $\texttt{Actor}_\texttt{gt}$ make decisions separately. \texttt{Critic} then selects the best action as its output. Upon further exploration, the agent detects more objects, and the \texttt{Adapter} adjusts its output, recognizing the task as stacking a ladle onto a plate and then placing them into a cabinet. The \texttt{Actors} and \texttt{Critic} subsequently make decisions based on the revised predictions.
\vspace{-0.1cm}}
\label{fig:demo3}
\end{figure}

At time step $t$, the planner receives an observed object list $y_t$ from observation functions \citep{HLSM}. We then apply the Adaptation module to predict the latent PDDL arguments $P$ based on $y_t$. The whole state $x_t$ is constructed by $y_t$ and $P$. With $x_t$, we invoke the actor-critic task planner \texttt{LLM-PL} to generate a future trajectory over a long horizon and return the sub-goal $S_k$. To ensure the output from the planner meets the requirements, a frozen BERT \citep{BERT} is used to map the output to the legal space. The proposed sub-goal will be executed by a low-level controller \texttt{Low-PL} \citep{HLSM}. When a sub-goal is completed or fails, the planner reinvokes the reasoning process to replan another future trajectory from the new state. The complete algorithm is presented in \Cref{alg:llm}, and \Cref{fig:demo3} provides an example of the entire process.

\subsection{Prompt Design}
\label{prompt_design}
All components follow a similar design. The prompt begins with an intuitive explanation of the task and a role description of the LLM. A frozen BERT is then used as a kNN retriever, encoding the task description and selecting $K$ examples with the closest Euclidean distance from the sample pool as in-context samples \citep{LLMplanner}. Intuitively, the planner would make similar suboptimal actions in similar tasks. For instance, if in an in-context sample ``Place two spray bottles into the cabinet,'' the planner fails to open the cabinet when putting the second spray bottle into it. In the current task ``Putting two candles in a cabinet'', the planner would know to avoid a similar mistake. The detailed prompts for each process can be viewed in \Cref{prompt-detail}.
\subsection{Hindsight Method}
\label{hindsight}

In \Cref{sec:motivation}, we gain a coherent framework to describe previous hindsight methods. However, we find that such methods can lead to the policy $\pi$ being suboptimal, particularly when the number of samples is insufficient. To illustrate this better, we consider the optimization objective in  \Cref{eq:rl2}. It aims to learn a policy under different values of  $z$ where $z\sim p(z)$. During the collection phase, the agent's trajectory is usually suboptimal and random. Assume the distribution of $I(\tau) \sim q$. The training objective after relabeling is to train a policy $\hat{\pi}$ satisfies that 
\begin{align}
    \hat \pi = \arg\max_{\pi}\ \mathbb{E}_{\tau \sim P(\cdot|\pi,z), z\sim q(z)}[R_z(\tau)],
\end{align}
Define the $\pi^*$ as the oracle. It is easy to see that
\begin{align}
\label{eq:cmp}
    \mathbb{E}_{\tau \sim P(\cdot|\hat\pi,z), z\sim p(z)}[R_z(\tau)]<\mathbb{E}_{\tau \sim P(\cdot|\pi^*,z), z\sim p(z)}[R_z(\tau)],
\end{align}
as the distribution of $z$ is shifted from  $p$ to $q$.

Based on such discovery, we propose a new method of hindsight.  Assume that $\tau^*$ is the ground truth rollout from the oracle $\pi^*$, we can rewrite $z=I(\tau^*)$, \Cref{eq:im} then becomes
\begin{align}
\min_\pi \mathbb{E}_{ \tau \sim P(\cdot|\pi,z), z\sim p(z)}\left[ \text{KL}(I(\tau), I(\tau^*)) \right] \label{eq:im2}.
\end{align}
Our method utilizes LLMs to relabel $\hat{\tau}=\tau_T+\{a_{T},x_{T+1},y_{T+1},a_{T+2},\ldots\}$ in such a way that $I(\hat{\tau})= I(\tau^*)=z$. Thus, we minimize the divergence in \Cref{eq:im2} while keeping the distribution of $z$ unshifted. Intuitively, \Cref{eq:cmp} shows that relabeling $z$ alters the distribution of tasks that are truly relevant to our daily lives. This is especially crucial in the reasoning process of EIF.

In practice, our hindsight method consists of two main parts: the collection phase and the deployment phase. During the collection phase, the planner executes tasks and retrieves $ K $ examples from a small set of ground truth samples. At each task, the planner generates a possibly suboptimal trajectory $ \tau $ and relabels them. The algorithm is summarized in \Cref{alg:collection} of \Cref{more-alg}. During the deployment phase, the $\texttt{Actor}_\texttt{gt} $ is prompted with ground truth samples while the $\texttt{Actor}_\texttt{hind}$ and the $\texttt{Critic}$ are prompted with relabeled samples. Intuitively, we hope that the $\texttt{Actor}_\texttt{gt} $ can provide the correct action to complete the task along the shortest path. However, when an incorrect action\textemdash which is often unavoidable\textemdash is executed, the $ \texttt{Actor}_\texttt{hind}$ and the $\texttt{Critic} $ should be able to correct it. The relabeling process utilizes the reasoning ability of LLMs to fit suboptimal trajectories into correct rollouts. The CoT \citep{COT} method is utilized in the relabeling process. We first prompt the LLM to generate a \textit{Think} about the suboptimal rollout and then prompt it to complete the suboptimal rollout based on the \textit{Think}. A comparison of the hindsight method with the supervised methods is shown in \Cref{fig:demo1}, while the right half of \Cref{fig:demo2} illustrates an example of the relabeling process.

\subsection{Adaptation Module}
\label{adapt_model}
In a POMDP, the adaptation module is used to predict the latent variables from the observed environment $y_t$ \citep{pomdphindsight,kumar2021rmarapidmotoradaptation} and construct the whole state $x_t=(\texttt{Adapter}(y_t),y_t)$. In practice, we utilize an LLM as the adaptation module and set PDDL arguments as the prediction target for it. The input prompt for the adaptation module begins with an intuitive explanation of ALFRED, followed by several in-context samples. At the end of the prompt is the current task and the object list. At each step, the object list is updated as the agent explores the environment.

The output from the adaptation module varies depending on the task description. Inspired by PDDL \citep{PDDL,PDDL_in_LLM} of ALFRED, the adaptation module needs to predict the following arguments at each step: (1) \textit{object\_target}: The specific object to be interacted with during the task. (2) \textit{parent\_target}: The final place for the object in the task. (3) \textit{mrecep\_target}: The container or vessel necessary for the task. (4) \textit{toggle\_target}: The device that needs to be toggled in the task. (5) \textit{object\_state}: Indicates whether the target object needs to be cleaned, heated, or cooled. (6) \textit{object\_sliced}: Determines if the object must be sliced. (7) \textit{two\_object}: Specifies whether the task involves handling and placing two objects. The adaptation module predicts these arguments at each time before reasoning. Then, the arguments are processed into a specific format to assist the task planner to sense the environment better.

\subsection{Task Planner}
\label{task_planner}
We adopt an actor-critic planner \citep{RAFA}. At each time step $t$, the planner receives $ x_t $ from the environment and the adaptation module. We initiate two Actors: $ \texttt{Actor}_\texttt{gt} $ and $ \texttt{Actor}_\texttt{hind} $, with different samples from the sample pool $ \cD $. For each state, we prompt each Actor to generate $ \frac{W}{2} $ actions. The \texttt{Critic} then selects the top $ B $ actions. A generator $ \psi $ generates the next state based on each action. In this way, we map $\texttt{Actors}$ and $\texttt{Critic}$ to $ B $ future trajectories and select the best future trajectory $ (x_t, a_t^*, \ldots) $ through \texttt{Critic}. $ a_t^* $ is then returned as the sub-goal for \texttt{Low-PL}. The left half of \Cref{fig:demo2} shows the reasoning process of the planner.



%% file: experiment.tex
\section{Experiment}

\subsection{Setups}
\label{implement_detail}
We validate our framework using the ALFRED benchmark \citep{ALFRED}. This benchmark assesses the agent's capability to execute a series of actions for long-horizon household tasks based on natural language task descriptions and egocentric vision. The ALFRED dataset consists of 25k annotations, 108 distinct objects, 7 types of tasks, and 120 scenes. The dataset is divided into training, validation, and testing splits. The validation and test splits contain ``seen'' subsets, which are part of the training fold, and ``unseen'' subsets, which are distinct from it. The evaluation is based on Success Rate (SR) and Goal Condition (GC). Given the inherent noise in natural language instructions and the complexities of long-horizon task planning, the ALFRED benchmark presents significant challenges for embodied agents in formulating robust and precise plans.

Similar to previous work \citep{LLMplanner,Socraticplanner}, we only utilize a few examples from the 21k training set annotations. For each of the 7 task types, we randomly select 20 trajectories as the initial sample pool. At the collection phase, we run our planner on the 140 trajectories and collect sub-optimal trajectories. During collection, the same task is not included as in-context samples.

We then give a detailed discussion of the relabeling process. Directly applying the task description from ALFRED may lead to unsatisfactory results, as the task description is often vague. For example, the task ``Put a chilled potato on the small black table'' requires the planner to put the \textit{potato} on a \textit{SideTable}. If the task description is applied directly, LLMs might focus incorrectly on the \textit{Black Table} and return an incorrect action ``PutObject BlackTable''. If the task description is not included in the prompt, it could lead LLMs to imitate the ground truth trajectory. However, planners usually have multiple ways to complete a certain task. For instance, in a task requiring the planner to slice an apple, after slicing the apple, the planner could put the \textit{Knife} on the \textit{DiningTable} or \textit{CounterTop}. To address this issue, we relabel the task based on the latent PDDL arguments. The task description ``Put a chilled potato on the small black table'' becomes ``Pick up one cooled potato and put it on the SideTable''. This approach helps clarify the task for the planner and reduces the ambiguity in instructions.

For the kNN retriever, we use a frozen BERT from \cite{huggingface}. We employ GPT-4 Turbo \citep{gpt4} as the target LLM and set temperature to 0. For the \texttt{Adapter}, 5 in-context examples are retrieved from the sample pool through the kNN retriever. For the \texttt{Actors} and \texttt{Critic} modules, 2 in-context examples are retrieved. The task planner uses beam search with a depth and width of 2. To preserve the few-shot assumption and ensure a fair comparison, we directly adopt the pretrained modules for navigation, perception, and low-level control from HLSM \citep{HLSM}.

\subsection{Main Results}
\begin{table}[tb]
\centering
\begin{tabular}{@{}lcccccccccccccccc@{}}
\toprule
\multirow{2}{*}{Model} & \multirow{2}{*} &\multicolumn{2}{c}{Test Seen} & \multicolumn{2}{c}{Test Unseen} \\ \cmidrule(l){3-4}  \cmidrule(l){5-6} 

& n-shot  & SR & GC & SR & GC  \\ \midrule
HiTUT~\citep{HITUT}&full& 13.63& 21.11& 11.12& 17.89\\
HLSM~\citep{HLSM}&full& 25.11& \underline{35.79}& \underline{20.27}& \underline{27.24}\\
FILM~\citep{FILM}&full& \underline{28.83}& \textbf{{39.55}}& \textbf{{27.80}}& \textbf{38.52}\\
MCR-Agent~\citep{MCR} &full &\textbf{30.13}& - &17.04&-\\\midrule
FILM (low inst.)~\citep{FILM}  &few &0.00 &4.23&0.20&6.71\\
LLM-Planner~\citep{LLMplanner}&few& {15.33}&{24.57}& {13.41}&{22.8} \\
LLM-Planner (low inst.)~\citep{LLMplanner}&few& \underline{18.80}&\underline{26.77}& \underline{16.42}&\underline{23.37} \\
Socratic-Planner\citep{Socraticplanner}&few&  13.24& 21.51& 
10.66&19.53\\
Hindsight-Planner (ours) &few &\textbf{ 25.51} &\textbf{34.74}& \textbf{18.77}& \textbf{28.29}\\\midrule
\bottomrule
\end{tabular}
\setlength{\belowcaptionskip}{-10pt}
\caption{\textbf{Comparison with the state-of-the-art methods on SR and GC in the test set.} Bold numbers represent the highest level of accuracy, whereas underlined numbers signify the second-highest accuracy for each experimental configuration. ``low inst.'' refers to the use of step-by-step instructions. }
\label{tab:1}
\end{table}

\begin{table}[tb]
\centering
\setlength{\tabcolsep}{2pt}
\begin{tabular}{@{}lccccccccc@{}}
\toprule
\multirow{2}{*}{Model} & \multirow{2}{*} & \multicolumn{2}{c}{Valid Seen} & \multicolumn{2}{c}{Valid Unseen} & \multicolumn{2}{c}{Test Seen}& \multicolumn{2}{c}{Test Unseen}\\ \cmidrule(l){3-4}  \cmidrule(l){5-6} \cmidrule(l){7-8} \cmidrule(l){9-10} 

& n-shot  & SR & GC & SR & GC   & SR & GC   & SR & GC  \\ \midrule

HLSM~\citep{HLSM}  &full& \textbf{29.63}&\textbf{38.74} &\underline{18.28} &\textbf{31.24} &\underline{25.11}&\textbf{ 35.79}&\textbf{ 20.27}& \underline{27.24}\\
LLM-Planner~\citep{LLMplanner}&few& 13.53&28.28& 12.92&25.35 &15.33&24.57& 13.41&22.8\\
Socratic-Planner~\citep{Socraticplanner} &few& 14.88 & 25.47& 13.40& 24.91 &13.24& 21.51& 
10.66&19.53\\
Hindsight-Planner (ours) &few & \underline{25.61} &\underline{34.95}& \textbf{19.00}& \underline{29.90}&\textbf{ 25.51} &\underline{34.74}& \underline{18.77}& \textbf{28.29}\\\midrule
\bottomrule
\end{tabular}
\vspace{-0.1cm}
\caption{\textbf{Comparison with the same lower-controller. }   Bold numbers represent the highest level of accuracy, whereas underlined numbers signify the second-highest accuracy for each experimental configuration.}
\label{tab:2}
\end{table}
We initially compare our method to other few-shot methods, as shown in \Cref{tab:1}. It is evident that our method achieves a $10.18$ and $5.36$ higher success rate in ``Test Seen'' and ``Test Unseen'' categories, respectively, compared to the previous state-of-the-art method (LLM-Planner) that uses high-level instructions only. Moreover, even when compared to methods utilizing low-level, step-by-step instructions, our method still demonstrates superior performance.

We also compare our method to the other approaches under the same low-level controller \citep{HLSM} in \Cref{tab:2}. The results indicate that our method not only significantly outperforms previous few-shot LLM planners but also, for the first time, a few-shot LLM method (with around 100 examples) nearly matches and even surpasses (SR in ``Valid Unseen'', ``Test Seen'', and GC in ``Test Unseen'') fully supervised (around 21k samples) methods. 
\subsection{Ablation Study}
\begin{table}[tb]
\centering
\begin{tabular}{@{}lccccccc@{}}
\toprule
 Task Type &Examine & Pick&Clean &Stack&Pick Two&Heat&Cool \\ \midrule
Base Method&40.42&50&15.18&9.56&30.65&7.48&21.43\\
W.O. Hindsight Method&39.36&49.29&16.96&7.82&29.84&7.47&10.31\\
W.O. Adaptation Module&35.1&47.1&8.93&6.09&32.25&9.34&18.26\\\midrule
\bottomrule
\end{tabular}
\setlength{\belowcaptionskip}{-10pt}
\caption{Ablation study on the success rate of different type of  tasks in  ``Valid Seen'' split.}
\label{tab:5}
\end{table}

\begin{wraptable}{r}{9.3cm}
\small
\vspace{-0.4cm}
\begin{tabular}{@{}lcccccccccccccccc@{}}
\toprule
\multirow{2}{*}{Model}  &\multicolumn{2}{c}{Valid Seen} & \multicolumn{2}{c}{Valid Unseen} \\ \cmidrule(l){2-3}  \cmidrule(l){4-5} 
 & SR & GC & SR & GC  \\ \midrule
 W.O. Adaptation Module &23.17 &33.28& 14.99& 27.36\\
 W.O. Hindsight Method& {23.53}&{32.76}& {16.32}&{28.06} \\
Base Method & {25.61}&{34.95}& {19.00}&{29.90} \\\midrule
\bottomrule
\end{tabular}
\setlength{\belowcaptionskip}{-10pt}
\caption{Ablation on ``Valid Seen'', ``Valid Unseen'' splits.}
\label{tab:3}
\end{wraptable}
We conduct ablation studies to understand the effectiveness of the components in our framework. First, we ablate the adaptation module \texttt{Adapter}, which requires the planner to make decisions based solely on the partially observed information. The results show that this causes a drop of $-2.44$ and $-4.01$ in the success rates for the ``Valid Seen'' and ``Valid Unseen'' splits. Then, we remove the hindsight prompts. For a fair comparison, the original planner requires both $\texttt{Actor}_\texttt{gt}$ and $\texttt{Actor}_\texttt{hind}$ to generate one action per state. We also ablate by prompting $\texttt{Actor}_\texttt{gt}$ to output two actions for each state. \Cref{tab:3} shows that the success rates drop by $-2.08$ and $-2.68$ in the ``Valid Seen'' and ``Valid Unseen'' splits. 




For a more comprehensive analysis, we report the success rates for each task type in the ``Valid Seen'' split, as shown in \Cref{tab:5}. Additionally, we also present the average sub-goal lengths in \Cref{tab:4}. This analysis reveals that hindsight prompting is especially crucial in relatively long-horizon tasks, such as ``Cool Object'' and ``Heat Object''. This is likely because, in long-horizon tasks, planners are more likely to output suboptimal actions, allowing the hindsight actor to correct its mistakes. 
\begin{wraptable}{r}{5.2cm}
\centering
\small
\vspace{-0.3cm}
\begin{tabular}{@{}lc@{}}
\toprule
 Task Type & Avg. Sub-Goal Len.\\ \midrule
 Examine& 2.07\\
  Pick& 2.48\\
 Pick Two& 5.70\\
 Stack& 5.63\\
 Clean& 7.25\\
 Cool& 10.36\\
 Heat& 12.78\\\midrule
\bottomrule
 
\end{tabular}
\caption{Average sub-goal lengths.}
\label{tab:4}
\end{wraptable}On the other hand, the adaptation module can assist the planner in better sensing the environment, leading to a general improvement across nearly all areas.

%% file: conclusion.tex
\section{Conclusion}
This paper explores an effective few-shot framework for Embodied Instruction Following. We approach the task as a POMDP and design a closed-loop Hindsight Planner equipped with an adaptation module to enhance the agent's environmental sensing capabilities. Compared to previous open-loop, supervised methods, our approach is more robust and performs better. Furthermore, the planner incorporates a novel hindsight method that enables it to learn from suboptimal trajectories. we hope our work inspires future research in this area.

%% file: appendix.tex
\section{More Algorithm}
\label{more-alg}
In \Cref{alg: example}, we present a beam search example of a hindsight planner. During the collection phase, one $\texttt{Actor}$ prompted from the ground truth sample pool is required to output $W$ actions for each state, and $\texttt{Critic}$ is used to retain the best $B$ actions for the next round of planning. When the search depth $U$ is reached, the best rollout is selected and the first action from it is returned. At the deployment phase, two $\texttt{Actors}$ are prompted with hindsight prompts and ground truth samples. Each $\texttt{Actor}$ is required to generate $\frac{W}{2}$ actions.

\Cref{alg:collection} outlines the algorithm for the collection phase. To preserve the few-shot assumption, the planner collects suboptimal trajectories from $\cD_\text{gt}$. During the execution of the current task, this task is specifically excluded from being used as an ICL sample to the planner. We employ a prompt generator $\phi$ to relabel tasks and mitigate ambiguity in the instructions.
\begin{algorithm}[h]\small
	\caption {LLM Planner: A Beam Search Example}
	\begin{algorithmic}[1]	\label{alg: example}
    \STATE \textbf{Input} $\texttt{Actors}$, $\texttt{Critic}$,  the initial state $s$, a generator $\psi$, the search
Breadth $B$, the proposal width $W$ and the search Depth $U$.

    \STATE  Set State $S_0\leftarrow \{s\}$.
    \STATE Set Action array $A_0\leftarrow \emptyset$.
    \STATE  Get numbers of $\texttt{Actors}$  $n\leftarrow $ $\texttt{len} ~(\texttt{Actors}) $.
    \FOR{ $u=0,\ldots, U$}
        \FOR{ $\texttt{Actor}_\texttt{i}$ in $\texttt{Actors}$}
            \STATE For each $s_u$ in $S_u$, invoke $\texttt{Actor}_\texttt{i}$ to propose  $\frac{W}{n}$ candidate actions.
        \ENDFOR
        \STATE For each $a_u^{(w)}$ invoke $\psi$ to generate next state $s_{u+1}^{(w)}$.
        \STATE For each tuple $(s_u,a_u^{(w)},s_{u+1}^{(w)})$, use \texttt{Critic} to evaluate the expected cumulative reward $V_{u+1}^{(w)}$.
        \STATE Select $B$ best  $(s_u,a_u^{(w)},s_{u+1}^{(w)})$ with highest $V$ and write them to $S_u\times A_u\times S_{u+1}$.
    \ENDFOR
    \STATE For $B$ preserved rollouts in $S_0\times A_0\times \ldots \times S_{U+1}$, invoke \texttt{Critic} to evaluate the expected cumulative reward $V_{u+1}^{(b)}$.
    \STATE Select the best rollout $(s_0^*,a_0^*,\ldots,s_{U+1}^*)$ and return $a_0^*$.
	\end{algorithmic}
\end{algorithm}

\begin{algorithm}[h]\small
	\caption {Hindsight Prompt}
	\begin{algorithmic}[1]	\label{alg:collection}
    \STATE \textbf{Input}: A ground truth sample pool $\cD_{gt}$, a prompt generator $\phi$.
    \STATE \textbf{Initialize} Initialize $\texttt{Agent}$ from $\cD_\text{gt}$, set $\cD_\text{hind}\leftarrow \emptyset$.
    \FOR{sample $s$ in $\cD_\text{gt}$}
        \STATE Extract ground truth rollout $R$, task description $I$, PDDL arguments $P$ from $s$.
        \STATE Initialize environment $E$ with $s$.
        \STATE Collect suboptimal trajectories ${traj}\leftarrow$ $\texttt{Agent}(I,E,D_\text{gt}/\{s\})$ (e.g. \Cref{alg: example} of \Cref{more-alg}).
        \STATE Rename the task description $\Tilde{I}\leftarrow \phi(P)$.
        \STATE Get reflection $\text{Think}\leftarrow \texttt{LLM}(\Tilde{I},\text{traj},R)$.
        \STATE Relabel trajectory $\text{prompt}_\text{actor}\leftarrow \texttt{LLM}(\Tilde{I},\text{traj},R,\text{Think})$.
        \STATE  Generate critic from suboptimal trajectory $\text{prompt}_\text{critic}\leftarrow \texttt{LLM}(\Tilde{I},\text{traj},R)$.
        \STATE Append $\text{prompt}_\text{actor},~\text{prompt}_\text{critic}$ to $\cD_\text{hind}$.
    \ENDFOR
    \STATE Build hindsight sample pool $\cD=\cD_\text{gt}\bigcup \cD_\text{hind}$.
    \STATE Initialize $\texttt{Actor}_\texttt{gt}$, $\texttt{Adapter}$ from $\cD_{gt}$,
    initial $\texttt{Critic}$, $\texttt{Actor}_\texttt{hind}$ from $\cD_\text{hind}$.
    \STATE  \textbf{Return} $\texttt{Actor}_{\theta}$, $\texttt{Critic}$, \texttt{Adapter}   for any $~\theta\in \{\texttt{gt,hind}\}$.
	\end{algorithmic}
\end{algorithm}
\input{prompt-main}

%% file: prompt-main.tex
\section{Prompts}
\label{prompt-detail}
\subsection{Prompts for Planner}
Here, we display prompts for various components. The \texttt{<base\_info>} defines the role descriptions while the \texttt{<samples>} provide in-context examples for the \texttt{Actors}, the\texttt{Critic}, and the\texttt{Adapter}.

We first show the role description for the \texttt{Actors}, the \texttt{Critic}, and the \texttt{Adapter}.
\vspace{0.3cm}
\begin{Verbatim}
Interact with a household to solve a task. 
At each step, you will be provided with the previous observations and action pairs.

Important: You **are required** to return an action.

The answer should contain two parts, the action type and a target. 

The allowed types of actions are: 

OpenObject, CloseObject, PickupObject, PutObject, ToggleObjectOn, ToggleObjectOff, SliceObject, Stop

The target of OpenObject, CloseObject, PickupObject, ToggleObjectOn, ToggleObjectOff, SliceObject is the object agent interacts with, and the target of PutObjectis the place to put the object.

Stop should end with NIL.Note if all requirements are satisfied, you just need to output Stop
\end{Verbatim}
\vspace{0.3cm}
\vspace{0.3cm}
\begin{Verbatim}
You are a value critic of states in a household task. You would be given a task description, some observations and actions, you need to give a critic about them. **Note Your critic should end with format: the value is a/b=...**

The allowed types of actions are: OpenObject, CloseObject, PickupObject, PutObject, ToggleObjectOn, ToggleObjectOff, SliceObject,Stop

The target of OpenObject, CloseObject, PickupObject, ToggleObjectOn, ToggleObjectOff, SliceObject is the object agent interacts with and the target of PutObjectis the place to put the object.

Explore and Stop should be followed with NIL.Note if all requirements are satisfied, you just need to output Stop. You might need to OpenObject so you can see the object you need to interact with.
\end{Verbatim}
\vspace{0.3cm}

\vspace{0.3cm}
\begin{Verbatim}
Predict the necessary components for the following household task:
-**Moveable Receptacle (mrecep_target)**: Identify any container or vessel required for the task. Return `None` if not applicable.
-**Object Slicing (object_sliced)**: Determine if the object needs to be sliced. Provide a boolean value (`True` for yes, `False` for no).
-**Object Target (object_target)**: Identify the specific object that is the focus of the task and will be interacted with. This could be the item that needs to be moved, cleaned, heated, cooled, sliced or examined.
-**Parent Target (parent_target)**: Specify the final resting place for the object or its parts. Return `None` if there is no designated location.
-**Toggle Target (toggle_target)**: Indicate any appliance or device that must be toggled during the task. Return `None` if no toggling is required.
-**Object State (object_state)**: Indicate whether the target object needs to be clean, heat, or cool. Return 'None' if no such action is required.
-**Two Objects (two_object)**: Specify whether the task requires the agent to handle and place two *identical* objects into the parent target location. Set to True if needed, otherwise False.  Note that this parameter should be True only when the task demands picking and placing two of the *same* items.
-**Note that the objects you need to predict might not been seen yet.
\end{Verbatim}
\vspace{0.3cm}

We then show the \texttt{<samples>} for the \texttt{Actors}, the \texttt{Critic}, and the \texttt{Adapter}. Since there are 140 samples for each component, we select just 2 samples from each to demonstrate.
\vspace{0.3cm}
\begin{Verbatim}
Task: Place a cup in the coffee maker.
The objects you seen are: Bread,ButterKnife,Cabinet,Chair,CoffeeMachine,CounterTop,Cup,DishSponge,Drawer,Fork,Fridge,GarbageCan,Lettuce,Microwave,Mirror,Mug,Pan,Plate,Pot,SaltShaker,Sink,SoapBottle,Spatula,Spoon,StoveBurner,StoveKnob,DiningTable,SideTable,Toaster,Window
Predict: 
mrecep_target: None
object_sliced: False
object_target: Mug
parent_target: CoffeeMachine
toggle_target: None
object_state: cool
two_object: False
Task: Warm a cup to make coffee
The objects you seen are: Apple,Bread,ButterKnife,Cabinet,CoffeeMachine,CounterTop,Cup,Drawer,Egg,Fork,Fridge,GarbageCan,HousePlant,Kettle,Knife,Ladle,Lettuce,Microwave,Mirror,Pan,PepperShaker,Pot,Potato,SaltShaker,Sink,Spatula,StoveBurner,StoveKnob,Toaster,Tomato,Window
Predict: 
mrecep_target: None
object_sliced: False
object_target: Mug
parent_target: CoffeeMachine
toggle_target: None
object_state: heat
two_object: False
\end{Verbatim}
\vspace{0.3cm}

\vspace{0.3cm}
\begin{Verbatim}
Task:Place a cup in the coffee maker.
The objects you have seen are:Bread,ButterKnife,Cabinet,Chair,CoffeeMachine,CounterTop,Cup,DishSponge,Drawer,Fork,Fridge,GarbageCan,Lettuce,Microwave,Mirror,Mug,Pan,Plate,Pot,SaltShaker,Sink,SoapBottle,Spatula,Spoon,StoveBurner,StoveKnob,DiningTable,SideTable,Toaster,Window
Act: OpenObject : Cabinet
>OK
Act: PickupObject : Mug
>OK
Act: CloseObject : Cabinet
>OK
Act: OpenObject : Fridge
>OK
Act: PutObject : Fridge
>OK
Act: CloseObject : Fridge
>OK
Act: OpenObject : Fridge
>OK
Act: PickupObject : Mug
>OK
Act: CloseObject : Fridge
>OK
Act: PutObject : CoffeeMachine
>OK
Act: Stop : NIL
>OK

Task:Pick up tissues, drop them in the tub.
The objects you have seen are:Bathtub,Cabinet,Candle,Cloth,CounterTop,Mirror,Plunger,ScrubBrush,Sink,SoapBar,SoapBottle,Toilet,ToiletPaper,ToiletPaperHanger,Towel,TowelHolder,Window
Act: PickupObject : Cloth
>OK
Act: PutObject : Bathtub
>OK
Act: Stop : NIL
>OK
\end{Verbatim}
\vspace{0.3cm}

\vspace{0.3cm}
\begin{Verbatim}
Task:Place a cup in the coffee maker.
The objects you have seen are:Bowl, Bread, ButterKnife, Cabinet, Chair, CoffeeMachine, CounterTop, Cup, DishSponge, Drawer, Fridge, GarbageCan, Lettuce, LightSwitch, Microwave, Mirror, Mug, Pan, PepperShaker, Plate, SaltShaker, Sink, SoapBottle, Spatula, Spoon, StoveBurner, StoveKnob, DiningTable, SideTable, Toaster, Window
Act: OpenObject : Cabinet
>OK
Act: PickupObject : Cup
>OK
Act: CloseObject : Cabinet
>OK
Act: PutObject : DiningTable
>OK
Act: OpenObject : Fridge
>OK
Act: PickupObject:Cup
>OK
Act: PutObject:Fridge
>OK
Act: CloseObject:Fridge
>OK
Act: OpenObject:Fridge
>OK
Act: PickupObject:Cup
>OK
Act: CloseObject:Fridge
>OK
Act: PutObject:CoffeeMachine
>OK
Act: Stop : NIL
>OK

Task:Pick up tissues, drop them in the tub.
The objects you have seen are:Bathtub, Cabinet, Candle, Cloth, CounterTop, GarbageCan, HandTowel, HandTowelHolder, LightSwitch, Mirror, Painting, ScrubBrush, Shelf, ShowerDoor, ShowerGlass, Sink, SoapBottle, Television, Toilet, ToiletPaperHanger, Towel, TowelHolder, Window
Act: PickupObject : TissueBox
>OK
Act: PutObject:CounterTop
>OK
Act: PickupObject:Cloth
>OK
Act: PutObject:Bathtub
>OK
Act: Stop : NIL
>OK
\end{Verbatim}
\vspace{0.3cm}

Having demonstrated the \texttt{<base\_info>} and \texttt{<samples>}, we can now present the prompt template for the \texttt{Actors}, the \texttt{Critic}, and the \texttt{Adapter}. Note that the prompt of $\texttt{Actor}_\texttt{gt}$ and $\texttt{Actor}_\texttt{hind}$ differ in \texttt{<samples>}. The \texttt{<object\_list>} indicates the objects the agent has seen in the environment. Meanwhile, the \texttt{<PDDL\_predicted>} represents to the output of the \texttt{Adapter}, and the \texttt{<K>} indicates the number of samples in each component. To facilitate better comprehension by LLMs, the PDDL arguments are converted into a natural language description. The \texttt{<previous\_history>} includes the previous actions executed by the agent, enabling the planner to make better decisions based on this information. Concurrently, a prompt generator reviews the \texttt{<previous\_history>} and outputs \texttt{<history\_information>} to assist LLMs in identifying objects being held and the open/closed status of containers.
\vspace{0.3cm}
\begin{Verbatim}
<Adapter_base_info>
Here are <K> examples:
<Adapter_samples>  
Your task is: <task_inst>
The objects you have seen are: <object_list>
\end{Verbatim}
\vspace{0.3cm}
\vspace{0.3cm}
\begin{Verbatim}
<Critic_base_info>
Here are <K> examples:
<Critic_samples>
Your task is: <task_inst>
Your knowledge about this task is: <PDDL_predicted>
The objects you have seen are: <object_list>
previous_history
Based on the **actions** and **Your knowledge about this task** , write a Critic.
Critic:
\end{Verbatim}
\vspace{0.3cm}
\vspace{0.3cm}
\begin{Verbatim}
<Actor_base_info>
Here are <K> examples:
<Actor_samples> 
Your task is: <task_inst>
Your knowledge about this task is: <PDDL_predicted>
The objects you have seen are: <object_list>
Your knowledge about the current state is: <history_information>
<previous_history>
Act: 
\end{Verbatim}
\vspace{0.3cm}
\subsection{Prompts for hindsight}
We now present the prompts used to query LLMs in our hindsight method. During the relabeling process for the \texttt{Actor}, we first prompt the LLMs to generate a \texttt{<Think>} for the suboptimal trajectory, and then query them to complete the task based on it. For the relabeling process of the \texttt{Critic}, we directly prompt the LLMs to generate an evaluation for the suboptimal trajectory. We first present the hindsight samples for clarity.
\vspace{0.3cm}
\begin{Verbatim}
Task: Put a fork on a table.
groundtruth rollout:
PickupObject:Fork
PutObject:Sink
ToggleObjectOn:Faucet
ToggleObjectOff:Fauce
PickupObject:Fork
PutObject:SideTable
Stop:NIL
The incomplete rollout:
PickupObject:Fork
PutObject:SideTable

Think: According to the groundtruth rollout, in this incomplete rollout, I don't clean the fork and the fork is on the sidetable, I need to pick up the fork and use faucet to clean the fork and put it onto the sidetable.
Task: Put a warmed apple in the fridge.
groundtruth rollout:
PickupObject:Apple
OpenObject:Microwave
PutObject:Microwave
CloseObject:Microwave
ToggleObjectOn:Microwave
ToggleObjectOff:Microwave
OpenObject:Microwave
PickupObject:Apple
CloseObject:Microwave
OpenObject:Fridge
PutObject:Fridge
CloseObject:Fridge
Stop:NIL
The incomplete rollout:
PickupObject : Apple
OpenObject : Fridge
PutObject : Fridge
CloseObject : Fridge
OpenObject : Fridge
PickupObject : Apple
CloseObject : Fridge
OpenObject : Microwave
Think: According to the groundtruth rollout, in this incomplete rollout, I don't heat the apple and the apple is in the fridge, I need to open the fridge, pickup the apple and use microwave to heat the apple, then I should put the apple back into the fridge.
\end{Verbatim}
\vspace{0.3cm}

\vspace{0.3cm}
\begin{Verbatim}
Task: Put a fork on a table.
groundtruth rollout:
PickupObject:Fork
PutObject:Sink
ToggleObjectOn:Faucet
ToggleObjectOff:Fauce
PickupObject:Fork
PutObject:SideTable
Stop:NIL
the incomplete rollout:
PickupObject:Fork
PutObject:SideTable

Think: According to the groundtruth rollout, in this incomplete rollout, I don't clean the fork and the fork is on the sidetable, I need to pick up the fork and use faucet to clean the fork and put it onto the sidetable.
Based on the Think and groundtruth rollout, the new actions append to the incomplete rollout are:
PickupObject : Fork
PutObject :Sink
ToggleObjectOn : Faucet
ToggleObjectOff : Faucet
PickupObject:  Fork
PutObject:SideTable
Stop : NIL

Task: Put a warmed apple in the fridge.
groundtruth rollout:
PickupObject : Apple
OpenObject : Microwave
PutObject : Microwave
CloseObject : Microwave
ToggleObjectOn : Microwave
ToggleObjectOff : Microwave
OpenObject : Microwave
PickupObject : Apple
CloseObject : Microwave
OpenObject : Fridge
PutObject : Fridge
CloseObject : Fridge
Stop:NIL
the incomplete rollout:
PickupObject : Apple
OpenObject : Fridge
PutObject : Fridge
CloseObject : Fridge
OpenObject : Fridge
PickupObject : Apple
CloseObject : Fridge
OpenObject : Microwave
Think: According to the groundtruth rollout, in this incomplete rollout, I don't heat the apple and the apple is in the fridge, I need to open the fridge, pickup the apple and use microwave to heat the apple, then I should put the apple back into the fridge.
Based on the Think and groundtruth rollout, the new actions append to the incomplete rollout are:
OpenObject : Fridge
PickupObject : Apple
CloseObject : Fridge
PutObject: Microwave
CloseObject: Microwave
ToggleObjectOn: Microwave
ToggleObjectOff : Microwave
OpenObject : Microwave
PickupObject : Apple
CloseObject : Microwave
OpenObject : Fridge
PutObject : Fridge
CloseObject : Fridge
\end{Verbatim}
\vspace{0.3cm}
\vspace{0.3cm}
\begin{Verbatim}
Your task is: Put the cooked tomato on the round table
The rollout by agent is: OpenObject : Fridge
PickupObject : Tomato
CloseObject : Fridge
OpenObject : Microwave
The **ground truth rollout** is:
PickupObject:Tomato OpenObject:Microwave
PutObject:Microwave
CloseObject:Microwave
ToggleObjectOn:Microwave
ToggleObjectOff:Microwave 
OpenObject:Microwave
PickupObject:Tomato
CloseObject:Microwave
PutObject:DiningTable
Stop:NIL
Based on the **ground truth rollout** , write a critic
Critic:In this task, I need to do the following things in order: Pick the tomato and put it into microwave, use microwave to heat it,pick the tomato from microwave and put it onto the DiningTable.There are 5 subgoals in orde, I only achieved first of them, the value is 1/5=0.2.

Your task is: Put a chilled mug in the bottom cabinet closest to the fridge.
The rollout by agent is: 
PickupObject : Mug
OpenObject : Fridge
PutObject : Fridge
CloseObject : Fridge
OpenObject : Fridge
PickupObject : Mug
CloseObject : Fridge
PutObject : Cabinet
The **ground truth rollout** is:
PickupObject:Mug
OpenObject:Fridge
PutObject:Fridge
CloseObject:Fridge
OpenObject:Fridge
PickupObject:Mug
CloseObject:Fridge
OpenObject:Cabinet
PutObject:Cabinet
CloseObject:Cabinet
Stop:NIL
Based on the **ground truth rollout** , write a critic
Critic:In this task, I need to do the following things in order: pick the mug and put it into the fridge, pick the mug from the fridge and put the mug into the cabinet. There are 3 subgoals in all, I achieved 2 of them, this is because I don't open the cabinet, so I can't put the mug into it, the value is 2/3=0.66
\end{Verbatim}
\vspace{0.3cm}
We provide the prompts used for querying \texttt{Actors} and \texttt{Critic}, respectively. The \texttt{<relabeled\_task>} indicates the task rewritten based on its PDDL, as detailed in \Cref{implement_detail}. The \texttt{<gt\_rollout>} represents the ground truth rollout, while the \texttt{<suboptimal\_rollout>} denotes the rollout collected by our agent. 
\vspace{0.3cm}
\begin{Verbatim}
You are a housework agent, you will be given a task, a ground truth rollout to complete this task, and an incomplete rollout.
Your goal is to consider what action you need to append to the incomplete rollout to complete the task. 
Important: You should use your knowledge to judge what actions need to do based on the ground truth rollout and incomplete rollout. eg: if the agent forget to open the fridge, then the action of put object into fridge should be counted as failed, so you should open the fridge and put the object into the fridge.
Important: the openable object (fridge,mrcrowave...) are initially closed, so you need to open them before put object in it.
Important: You can hold one object in your hand at once.

The allowed types of actions are: OpenObject,CloseObject,PickupObject,PutObject,ToggleObjectOn,ToggleObjectOff,SliceObject,Stop
The target of actions like OpenObject, CloseObject, PickupObject, ToggleObjectOn, ToggleObjectOff, and SliceObject is the object the agent interacts with, whereas the target of PutObject is the location where the object is to be placed.
The 'Stop' action should be followed by 'NIL'. Note that if all requirements are met, you only need to output 'Stop'. Remember that you can only pick up one item at a time, so you must put down the object in your hand before picking up a new one.

Here are k examples:
<Actor_Think_samples>  
Task: <relabeled_task>
Ground truth rollout: <gt_rollout>
The incomplete rollout: <suboptimal_rollout>
Think:
\end{Verbatim}
\vspace{0.3cm}
\vspace{0.3cm}
\begin{Verbatim}
You are a housework agent, you will be given a task, a ground truth rollout to complete this task,  an incomplete rollout, and a think about the incomplete rollout.
Your goal is to finish the incomplete rollout based on the groundtruth rollout and your think. 
Important: You can only output the needed actions,seperated by '
', you must not output other things

The allowed types of actions are: OpenObject,CloseObject,PickupObject,PutObject,ToggleObjectOn,ToggleObjectOff,SliceObject,Stop
The target of actions like OpenObject, CloseObject, PickupObject, ToggleObjectOn, ToggleObjectOff, and SliceObject is the object the agent interacts with, whereas the target of PutObject is the location where the object is to be placed.
The 'Stop' action should be followed by 'NIL'. Note that if all requirements are met, you only need to output 'Stop'. Remember that you can only pick up one item at a time, so you must put down the object in your hand before picking up a new one.

Here are k examples:
<Actor_Complete_samples>
Task: <relabeled_task>
Ground truth rollout: <gt_rollout>
The incomplete  rollout: <suboptimal_rollout>
Think: <Think>
Based on the Think and groundtruth rollout, the new actions append to the incomplete rollout are:
\end{Verbatim}
\vspace{0.3cm}
\vspace{0.3cm}
\begin{Verbatim}
You will be provided with a household task roll-out conducted by an agent and a ground truth roll-out. Your task is to write a critic of the agent's roll-out based on the **ground truth rollout** The critic should follow the form:In this task, I need do the follwing things in order:... There are ... subgoals I need to achieve,My current state achieve ...
Important: You should use your knowledge to judge how many subgoals are achieved. eg: if the agent forget to open the fridge, then the action of put object into fridge should not counted.
Important: Your critic should end with "the value is a/b=.." You can round it into 2 decimal.
Important: You should write your critic based on given format, you should't output other things.
Important: You shouldn't mention about ground truth rollout in your critic.
Here are k examples: <Critic_samples>
Your task is: <relabelled_task>
The rollout by agent is:: <suboptimal_rollout>
The **ground truth rollout** is: <gt_rollout>
Based on the **ground truth rollout** , write a critic
Critic:
\end{Verbatim}
\vspace{0.3cm}

%% file: main.bbl
\begin{thebibliography}{39}
\expandafter\ifx\csname natexlab\endcsname\relax\def\natexlab#1{#1}\fi
\expandafter\ifx\csname url\endcsname\relax
  \def\url#1{\texttt{#1}}\fi
\expandafter\ifx\csname urlprefix\endcsname\relax\def\urlprefix{}\fi

\bibitem[{Ahn et~al.(2022)Ahn, Brohan, Brown, Chebotar, Cortes, David, Finn, Fu, Gopalakrishnan, Hausman, Herzog, Ho, Hsu, Ibarz, Ichter, Irpan, Jang, Ruano, Jeffrey, Jesmonth, Joshi, Julian, Kalashnikov, Kuang, Lee, Levine, Lu, Luu, Parada, Pastor, Quiambao, Rao, Rettinghouse, Reyes, Sermanet, Sievers, Tan, Toshev, Vanhoucke, Xia, Xiao, Xu, Xu, Yan and Zeng}]{ahn2022icanisay}
\text{Ahn, M.}, \text{Brohan, A.}, \text{Brown, N.}, \text{Chebotar, Y.}, \text{Cortes, O.}, \text{David, B.}, \text{Finn, C.}, \text{Fu, C.}, \text{Gopalakrishnan, K.}, \text{Hausman, K.}, \text{Herzog, A.}, \text{Ho, D.}, \text{Hsu, J.}, \text{Ibarz, J.}, \text{Ichter, B.}, \text{Irpan, A.}, \text{Jang, E.}, \text{Ruano, R.~J.}, \text{Jeffrey, K.}, \text{Jesmonth, S.}, \text{Joshi, N.~J.}, \text{Julian, R.}, \text{Kalashnikov, D.}, \text{Kuang, Y.}, \text{Lee, K.-H.}, \text{Levine, S.}, \text{Lu, Y.}, \text{Luu, L.}, \text{Parada, C.}, \text{Pastor, P.}, \text{Quiambao, J.}, \text{Rao, K.}, \text{Rettinghouse, J.}, \text{Reyes, D.}, \text{Sermanet, P.}, \text{Sievers, N.}, \text{Tan, C.}, \text{Toshev, A.}, \text{Vanhoucke, V.}, \text{Xia, F.}, \text{Xiao, T.}, \text{Xu, P.}, \text{Xu, S.}, \text{Yan, M.} and \text{Zeng, A.} (2022).
\newblock Do as i can, not as i say: Grounding language in robotic affordances.
\newline\urlprefix\url{https://arxiv.org/abs/2204.01691}

\bibitem[{Andrychowicz et~al.(2018)Andrychowicz, Wolski, Ray, Schneider, Fong, Welinder, McGrew, Tobin, Abbeel and Zaremba}]{andrychowicz2018hindsightexperiencereplay}
\text{Andrychowicz, M.}, \text{Wolski, F.}, \text{Ray, A.}, \text{Schneider, J.}, \text{Fong, R.}, \text{Welinder, P.}, \text{McGrew, B.}, \text{Tobin, J.}, \text{Abbeel, P.} and \text{Zaremba, W.} (2018).
\newblock Hindsight experience replay.
\newline\urlprefix\url{https://arxiv.org/abs/1707.01495}

\bibitem[{Bhambri et~al.(2024)Bhambri, Kim and Choi}]{MCR}
\text{Bhambri, S.}, \text{Kim, B.} and \text{Choi, J.} (2024).
\newblock Multi-level compositional reasoning for interactive instruction following.
\newline\urlprefix\url{https://arxiv.org/abs/2308.09387}

\bibitem[{Blukis et~al.(2021)Blukis, Paxton, Fox, Garg and Artzi}]{HLSM}
\text{Blukis, V.}, \text{Paxton, C.}, \text{Fox, D.}, \text{Garg, A.} and \text{Artzi, Y.} (2021).
\newblock A persistent spatial semantic representation for high-level natural language instruction execution.
\newblock \textit{Cornell University - arXiv,Cornell University - arXiv}.

\bibitem[{Brown et~al.(2020)Brown, Mann, Ryder, Subbiah, Kaplan, Dhariwal, Neelakantan, Shyam, Sastry, Askell, Agarwal, Herbert-Voss, Krueger, Henighan, Child, Ramesh, Ziegler, Wu, Winter, Hesse, Chen, Sigler, Litwin, Gray, Chess, Clark, Berner, McCandlish, Radford, Sutskever and Amodei}]{llm_few_shot_learner}
\text{Brown, T.~B.}, \text{Mann, B.}, \text{Ryder, N.}, \text{Subbiah, M.}, \text{Kaplan, J.}, \text{Dhariwal, P.}, \text{Neelakantan, A.}, \text{Shyam, P.}, \text{Sastry, G.}, \text{Askell, A.}, \text{Agarwal, S.}, \text{Herbert-Voss, A.}, \text{Krueger, G.}, \text{Henighan, T.}, \text{Child, R.}, \text{Ramesh, A.}, \text{Ziegler, D.~M.}, \text{Wu, J.}, \text{Winter, C.}, \text{Hesse, C.}, \text{Chen, M.}, \text{Sigler, E.}, \text{Litwin, M.}, \text{Gray, S.}, \text{Chess, B.}, \text{Clark, J.}, \text{Berner, C.}, \text{McCandlish, S.}, \text{Radford, A.}, \text{Sutskever, I.} and \text{Amodei, D.} (2020).
\newblock Language models are few-shot learners.
\newline\urlprefix\url{https://arxiv.org/abs/2005.14165}

\bibitem[{Chapman(1987)}]{PDDL}
\text{Chapman, D.} (1987).
\newblock Planning for conjunctive goals.
\newblock \textit{Artif. Intell.}, \textbf{32} 333--377.
\newline\urlprefix\url{https://api.semanticscholar.org/CorpusID:1525549}

\bibitem[{Dai et~al.(2023)Dai, Sun, Dong, Hao, Ma, Sui and Wei}]{dai2023gptlearnincontextlanguage}
\text{Dai, D.}, \text{Sun, Y.}, \text{Dong, L.}, \text{Hao, Y.}, \text{Ma, S.}, \text{Sui, Z.} and \text{Wei, F.} (2023).
\newblock Why can gpt learn in-context? language models implicitly perform gradient descent as meta-optimizers.
\newline\urlprefix\url{https://arxiv.org/abs/2212.10559}

\bibitem[{Devlin et~al.(2019)Devlin, Chang, Lee and Toutanova}]{BERT}
\text{Devlin, J.}, \text{Chang, M.-W.}, \text{Lee, K.} and \text{Toutanova, K.} (2019).
\newblock Bert: Pre-training of deep bidirectional transformers for language understanding.
\newline\urlprefix\url{https://arxiv.org/abs/1810.04805}

\bibitem[{Dong et~al.(2024)Dong, Li, Dai, Zheng, Ma, Li, Xia, Xu, Wu, Chang, Sun, Li and Sui}]{ICL_servey}
\text{Dong, Q.}, \text{Li, L.}, \text{Dai, D.}, \text{Zheng, C.}, \text{Ma, J.}, \text{Li, R.}, \text{Xia, H.}, \text{Xu, J.}, \text{Wu, Z.}, \text{Chang, B.}, \text{Sun, X.}, \text{Li, L.} and \text{Sui, Z.} (2024).
\newblock A survey on in-context learning.
\newline\urlprefix\url{https://arxiv.org/abs/2301.00234}

\bibitem[{Eysenbach et~al.(2020)Eysenbach, Geng, Levine and Salakhutdinov}]{eysenbach2020rewritinghistoryinverserl}
\text{Eysenbach, B.}, \text{Geng, X.}, \text{Levine, S.} and \text{Salakhutdinov, R.} (2020).
\newblock Rewriting history with inverse rl: Hindsight inference for policy improvement.
\newline\urlprefix\url{https://arxiv.org/abs/2002.11089}

\bibitem[{Furuta et~al.(2022)Furuta, Matsuo and Gu}]{furuta2022generalizeddecisiontransformeroffline}
\text{Furuta, H.}, \text{Matsuo, Y.} and \text{Gu, S.~S.} (2022).
\newblock Generalized decision transformer for offline hindsight information matching.
\newline\urlprefix\url{https://arxiv.org/abs/2111.10364}

\bibitem[{Ghosh et~al.(2019)Ghosh, Gupta, Fu, Reddy, Devin, Eysenbach and Levine}]{Ghosh2019LearningTR}
\text{Ghosh, D.}, \text{Gupta, A.}, \text{Fu, J.}, \text{Reddy, A.}, \text{Devin, C.}, \text{Eysenbach, B.} and \text{Levine, S.} (2019).
\newblock Learning to reach goals without reinforcement learning.
\newblock \textit{ArXiv}, \textbf{abs/1912.06088}.

\bibitem[{Guo et~al.(2021)Guo, Zhang, Zhang, Peng, Yi, Du, Hu, Guo and Chen}]{guo2021hindsightvaluefunctionvariance}
\text{Guo, J.}, \text{Zhang, R.}, \text{Zhang, X.}, \text{Peng, S.}, \text{Yi, Q.}, \text{Du, Z.}, \text{Hu, X.}, \text{Guo, Q.} and \text{Chen, Y.} (2021).
\newblock Hindsight value function for variance reduction in stochastic dynamic environment.
\newline\urlprefix\url{https://arxiv.org/abs/2107.12216}

\bibitem[{Hazan et~al.(2019)Hazan, Kakade, Singh and Soest}]{hazan2019provablyefficientmaximumentropy}
\text{Hazan, E.}, \text{Kakade, S.~M.}, \text{Singh, K.} and \text{Soest, A.~V.} (2019).
\newblock Provably efficient maximum entropy exploration.
\newline\urlprefix\url{https://arxiv.org/abs/1812.02690}

\bibitem[{Kim et~al.(2024)Kim, Kim, Kim, Min and Choi}]{CAPEAN}
\text{Kim, B.}, \text{Kim, J.}, \text{Kim, Y.}, \text{Min, C.} and \text{Choi, J.} (2024).
\newblock Context-aware planning and environment-aware memory for instruction following embodied agents.
\newline\urlprefix\url{https://arxiv.org/abs/2308.07241}

\bibitem[{Kumar et~al.(2021)Kumar, Fu, Pathak and Malik}]{kumar2021rmarapidmotoradaptation}
\text{Kumar, A.}, \text{Fu, Z.}, \text{Pathak, D.} and \text{Malik, J.} (2021).
\newblock Rma: Rapid motor adaptation for legged robots.
\newline\urlprefix\url{https://arxiv.org/abs/2107.04034}

\bibitem[{Lee et~al.(2023)Lee, Agarwal, Dann and Zhang}]{pomdphindsight}
\text{Lee, J.~N.}, \text{Agarwal, A.}, \text{Dann, C.} and \text{Zhang, T.} (2023).
\newblock Learning in pomdps is sample-efficient with hindsight observability.
\newline\urlprefix\url{https://arxiv.org/abs/2301.13857}

\bibitem[{Lee et~al.(2020)Lee, Eysenbach, Parisotto, Xing, Levine and Salakhutdinov}]{SMM}
\text{Lee, L.}, \text{Eysenbach, B.}, \text{Parisotto, E.}, \text{Xing, E.}, \text{Levine, S.} and \text{Salakhutdinov, R.} (2020).
\newblock Efficient exploration via state marginal matching.
\newline\urlprefix\url{https://arxiv.org/abs/1906.05274}

\bibitem[{Li et~al.(2020)Li, Pinto and Abbeel}]{li2020generalizedhindsightreinforcementlearning}
\text{Li, A.~C.}, \text{Pinto, L.} and \text{Abbeel, P.} (2020).
\newblock Generalized hindsight for reinforcement learning.
\newline\urlprefix\url{https://arxiv.org/abs/2002.11708}

\bibitem[{Liu et~al.(2024)Liu, Hu, Zhang, Guo, Ke, Liu and Wang}]{RAFA}
\text{Liu, Z.}, \text{Hu, H.}, \text{Zhang, S.}, \text{Guo, H.}, \text{Ke, S.}, \text{Liu, B.} and \text{Wang, Z.} (2024).
\newblock Reason for future, act for now: A principled framework for autonomous llm agents with provable sample efficiency.
\newline\urlprefix\url{https://arxiv.org/abs/2309.17382}

\bibitem[{Min et~al.(2021)Min, Chaplot, Ravikumar, Bisk and Salakhutdinov}]{FILM}
\text{Min, S.}, \text{Chaplot, D.}, \text{Ravikumar, P.}, \text{Bisk, Y.} and \text{Salakhutdinov, R.} (2021).
\newblock Film: Following instructions in language with modular methods.
\newblock \textit{Learning,Learning}.

\bibitem[{OpenAI et~al.(2024)OpenAI, Achiam, Adler, Agarwal, Ahmad, Akkaya, Aleman, Almeida, Altenschmidt, Altman, Anadkat, Avila, Babuschkin, Balaji, Balcom, Baltescu, Bao, Bavarian, Belgum, Bello, Berdine, Bernadett-Shapiro, Berner, Bogdonoff, Boiko, Boyd, Brakman, Brockman, Brooks, Brundage, Button, Cai, Campbell, Cann, Carey, Carlson, Carmichael, Chan, Chang, Chantzis, Chen, Chen, Chen, Chen, Chen, Chess, Cho, Chu, Chung, Cummings, Currier, Dai, Decareaux, Degry, Deutsch, Deville, Dhar, Dohan, Dowling, Dunning, Ecoffet, Eleti, Eloundou, Farhi, Fedus, Felix, Fishman, Forte, Fulford, Gao, Georges, Gibson, Goel, Gogineni, Goh, Gontijo-Lopes, Gordon, Grafstein, Gray, Greene, Gross, Gu, Guo, Hallacy, Han, Harris, He, Heaton, Heidecke, Hesse, Hickey, Hickey, Hoeschele, Houghton, Hsu, Hu, Hu, Huizinga, Jain, Jain, Jang, Jiang, Jiang, Jin, Jin, Jomoto, Jonn, Jun, Kaftan, Łukasz Kaiser, Kamali, Kanitscheider, Keskar, Khan, Kilpatrick, Kim, Kim, Kim, Kirchner, Kiros, Knight, Kokotajlo, Łukasz Kondraciuk,
  Kondrich, Konstantinidis, Kosic, Krueger, Kuo, Lampe, Lan, Lee, Leike, Leung, Levy, Li, Lim, Lin, Lin, Litwin, Lopez, Lowe, Lue, Makanju, Malfacini, Manning, Markov, Markovski, Martin, Mayer, Mayne, McGrew, McKinney, McLeavey, McMillan, McNeil, Medina, Mehta, Menick, Metz, Mishchenko, Mishkin, Monaco, Morikawa, Mossing, Mu, Murati, Murk, Mély, Nair, Nakano, Nayak, Neelakantan, Ngo, Noh, Ouyang, O'Keefe, Pachocki, Paino, Palermo, Pantuliano, Parascandolo, Parish, Parparita, Passos, Pavlov, Peng, Perelman, de~Avila Belbute~Peres, Petrov, de~Oliveira~Pinto, Michael, Pokorny, Pokrass, Pong, Powell, Power, Power, Proehl, Puri, Radford, Rae, Ramesh, Raymond, Real, Rimbach, Ross, Rotsted, Roussez, Ryder, Saltarelli, Sanders, Santurkar, Sastry, Schmidt, Schnurr, Schulman, Selsam, Sheppard, Sherbakov, Shieh, Shoker, Shyam, Sidor, Sigler, Simens, Sitkin, Slama, Sohl, Sokolowsky, Song, Staudacher, Such, Summers, Sutskever, Tang, Tezak, Thompson, Tillet, Tootoonchian, Tseng, Tuggle, Turley, Tworek, Uribe, Vallone,
  Vijayvergiya, Voss, Wainwright, Wang, Wang, Wang, Ward, Wei, Weinmann, Welihinda, Welinder, Weng, Weng, Wiethoff, Willner, Winter, Wolrich, Wong, Workman, Wu, Wu, Wu, Xiao, Xu, Yoo, Yu, Yuan, Zaremba, Zellers, Zhang, Zhang, Zhao, Zheng, Zhuang, Zhuk and Zoph}]{gpt4}
\text{OpenAI}, \text{Achiam, J.}, \text{Adler, S.}, \text{Agarwal, S.}, \text{Ahmad, L.}, \text{Akkaya, I.}, \text{Aleman, F.~L.}, \text{Almeida, D.}, \text{Altenschmidt, J.}, \text{Altman, S.}, \text{Anadkat, S.}, \text{Avila, R.}, \text{Babuschkin, I.}, \text{Balaji, S.}, \text{Balcom, V.}, \text{Baltescu, P.}, \text{Bao, H.}, \text{Bavarian, M.}, \text{Belgum, J.}, \text{Bello, I.}, \text{Berdine, J.}, \text{Bernadett-Shapiro, G.}, \text{Berner, C.}, \text{Bogdonoff, L.}, \text{Boiko, O.}, \text{Boyd, M.}, \text{Brakman, A.-L.}, \text{Brockman, G.}, \text{Brooks, T.}, \text{Brundage, M.}, \text{Button, K.}, \text{Cai, T.}, \text{Campbell, R.}, \text{Cann, A.}, \text{Carey, B.}, \text{Carlson, C.}, \text{Carmichael, R.}, \text{Chan, B.}, \text{Chang, C.}, \text{Chantzis, F.}, \text{Chen, D.}, \text{Chen, S.}, \text{Chen, R.}, \text{Chen, J.}, \text{Chen, M.}, \text{Chess, B.}, \text{Cho, C.}, \text{Chu, C.}, \text{Chung, H.~W.}, \text{Cummings, D.}, \text{Currier, J.}, \text{Dai, Y.}, \text{Decareaux, C.},
  \text{Degry, T.}, \text{Deutsch, N.}, \text{Deville, D.}, \text{Dhar, A.}, \text{Dohan, D.}, \text{Dowling, S.}, \text{Dunning, S.}, \text{Ecoffet, A.}, \text{Eleti, A.}, \text{Eloundou, T.}, \text{Farhi, D.}, \text{Fedus, L.}, \text{Felix, N.}, \text{Fishman, S.~P.}, \text{Forte, J.}, \text{Fulford, I.}, \text{Gao, L.}, \text{Georges, E.}, \text{Gibson, C.}, \text{Goel, V.}, \text{Gogineni, T.}, \text{Goh, G.}, \text{Gontijo-Lopes, R.}, \text{Gordon, J.}, \text{Grafstein, M.}, \text{Gray, S.}, \text{Greene, R.}, \text{Gross, J.}, \text{Gu, S.~S.}, \text{Guo, Y.}, \text{Hallacy, C.}, \text{Han, J.}, \text{Harris, J.}, \text{He, Y.}, \text{Heaton, M.}, \text{Heidecke, J.}, \text{Hesse, C.}, \text{Hickey, A.}, \text{Hickey, W.}, \text{Hoeschele, P.}, \text{Houghton, B.}, \text{Hsu, K.}, \text{Hu, S.}, \text{Hu, X.}, \text{Huizinga, J.}, \text{Jain, S.}, \text{Jain, S.}, \text{Jang, J.}, \text{Jiang, A.}, \text{Jiang, R.}, \text{Jin, H.}, \text{Jin, D.}, \text{Jomoto, S.}, \text{Jonn, B.}, \text{Jun, H.},
  \text{Kaftan, T.}, \text{Łukasz Kaiser}, \text{Kamali, A.}, \text{Kanitscheider, I.}, \text{Keskar, N.~S.}, \text{Khan, T.}, \text{Kilpatrick, L.}, \text{Kim, J.~W.}, \text{Kim, C.}, \text{Kim, Y.}, \text{Kirchner, J.~H.}, \text{Kiros, J.}, \text{Knight, M.}, \text{Kokotajlo, D.}, \text{Łukasz Kondraciuk}, \text{Kondrich, A.}, \text{Konstantinidis, A.}, \text{Kosic, K.}, \text{Krueger, G.}, \text{Kuo, V.}, \text{Lampe, M.}, \text{Lan, I.}, \text{Lee, T.}, \text{Leike, J.}, \text{Leung, J.}, \text{Levy, D.}, \text{Li, C.~M.}, \text{Lim, R.}, \text{Lin, M.}, \text{Lin, S.}, \text{Litwin, M.}, \text{Lopez, T.}, \text{Lowe, R.}, \text{Lue, P.}, \text{Makanju, A.}, \text{Malfacini, K.}, \text{Manning, S.}, \text{Markov, T.}, \text{Markovski, Y.}, \text{Martin, B.}, \text{Mayer, K.}, \text{Mayne, A.}, \text{McGrew, B.}, \text{McKinney, S.~M.}, \text{McLeavey, C.}, \text{McMillan, P.}, \text{McNeil, J.}, \text{Medina, D.}, \text{Mehta, A.}, \text{Menick, J.}, \text{Metz, L.}, \text{Mishchenko, A.},
  \text{Mishkin, P.}, \text{Monaco, V.}, \text{Morikawa, E.}, \text{Mossing, D.}, \text{Mu, T.}, \text{Murati, M.}, \text{Murk, O.}, \text{Mély, D.}, \text{Nair, A.}, \text{Nakano, R.}, \text{Nayak, R.}, \text{Neelakantan, A.}, \text{Ngo, R.}, \text{Noh, H.}, \text{Ouyang, L.}, \text{O'Keefe, C.}, \text{Pachocki, J.}, \text{Paino, A.}, \text{Palermo, J.}, \text{Pantuliano, A.}, \text{Parascandolo, G.}, \text{Parish, J.}, \text{Parparita, E.}, \text{Passos, A.}, \text{Pavlov, M.}, \text{Peng, A.}, \text{Perelman, A.}, \text{de~Avila Belbute~Peres, F.}, \text{Petrov, M.}, \text{de~Oliveira~Pinto, H.~P.}, \text{Michael}, \text{Pokorny}, \text{Pokrass, M.}, \text{Pong, V.~H.}, \text{Powell, T.}, \text{Power, A.}, \text{Power, B.}, \text{Proehl, E.}, \text{Puri, R.}, \text{Radford, A.}, \text{Rae, J.}, \text{Ramesh, A.}, \text{Raymond, C.}, \text{Real, F.}, \text{Rimbach, K.}, \text{Ross, C.}, \text{Rotsted, B.}, \text{Roussez, H.}, \text{Ryder, N.}, \text{Saltarelli, M.}, \text{Sanders, T.}, \text{Santurkar,
  S.}, \text{Sastry, G.}, \text{Schmidt, H.}, \text{Schnurr, D.}, \text{Schulman, J.}, \text{Selsam, D.}, \text{Sheppard, K.}, \text{Sherbakov, T.}, \text{Shieh, J.}, \text{Shoker, S.}, \text{Shyam, P.}, \text{Sidor, S.}, \text{Sigler, E.}, \text{Simens, M.}, \text{Sitkin, J.}, \text{Slama, K.}, \text{Sohl, I.}, \text{Sokolowsky, B.}, \text{Song, Y.}, \text{Staudacher, N.}, \text{Such, F.~P.}, \text{Summers, N.}, \text{Sutskever, I.}, \text{Tang, J.}, \text{Tezak, N.}, \text{Thompson, M.~B.}, \text{Tillet, P.}, \text{Tootoonchian, A.}, \text{Tseng, E.}, \text{Tuggle, P.}, \text{Turley, N.}, \text{Tworek, J.}, \text{Uribe, J. F.~C.}, \text{Vallone, A.}, \text{Vijayvergiya, A.}, \text{Voss, C.}, \text{Wainwright, C.}, \text{Wang, J.~J.}, \text{Wang, A.}, \text{Wang, B.}, \text{Ward, J.}, \text{Wei, J.}, \text{Weinmann, C.}, \text{Welihinda, A.}, \text{Welinder, P.}, \text{Weng, J.}, \text{Weng, L.}, \text{Wiethoff, M.}, \text{Willner, D.}, \text{Winter, C.}, \text{Wolrich, S.}, \text{Wong, H.}, \text{Workman,
  L.}, \text{Wu, S.}, \text{Wu, J.}, \text{Wu, M.}, \text{Xiao, K.}, \text{Xu, T.}, \text{Yoo, S.}, \text{Yu, K.}, \text{Yuan, Q.}, \text{Zaremba, W.}, \text{Zellers, R.}, \text{Zhang, C.}, \text{Zhang, M.}, \text{Zhao, S.}, \text{Zheng, T.}, \text{Zhuang, J.}, \text{Zhuk, W.} and \text{Zoph, B.} (2024).
\newblock Gpt-4 technical report.
\newline\urlprefix\url{https://arxiv.org/abs/2303.08774}

\bibitem[{Peng et~al.(2020)Peng, Coumans, Zhang, Lee, Tan and Levine}]{peng2020learningagileroboticlocomotion}
\text{Peng, X.~B.}, \text{Coumans, E.}, \text{Zhang, T.}, \text{Lee, T.-W.}, \text{Tan, J.} and \text{Levine, S.} (2020).
\newblock Learning agile robotic locomotion skills by imitating animals.
\newline\urlprefix\url{https://arxiv.org/abs/2004.00784}

\bibitem[{Pong et~al.(2020)Pong, Gu, Dalal and Levine}]{pong2020temporaldifferencemodelsmodelfree}
\text{Pong, V.}, \text{Gu, S.}, \text{Dalal, M.} and \text{Levine, S.} (2020).
\newblock Temporal difference models: Model-free deep rl for model-based control.
\newline\urlprefix\url{https://arxiv.org/abs/1802.09081}

\bibitem[{Shin et~al.(2024)Shin, jeon, Kim, Kang and Zhang}]{Socraticplanner}
\text{Shin, S.}, \text{jeon, S.}, \text{Kim, J.}, \text{Kang, G.-C.} and \text{Zhang, B.-T.} (2024).
\newblock Socratic planner: Inquiry-based zero-shot planning for embodied instruction following.
\newline\urlprefix\url{https://arxiv.org/abs/2404.15190}

\bibitem[{Shridhar et~al.(2020)Shridhar, Thomason, Gordon, Bisk, Han, Mottaghi, Zettlemoyer and Fox}]{ALFRED}
\text{Shridhar, M.}, \text{Thomason, J.}, \text{Gordon, D.}, \text{Bisk, Y.}, \text{Han, W.}, \text{Mottaghi, R.}, \text{Zettlemoyer, L.} and \text{Fox, D.} (2020).
\newblock Alfred: A benchmark for interpreting grounded instructions for everyday tasks.
\newblock In \textit{2020 IEEE/CVF Conference on Computer Vision and Pattern Recognition (CVPR)}.
\newline\urlprefix\url{http://dx.doi.org/10.1109/cvpr42600.2020.01075}

\bibitem[{Silver et~al.(2023)Silver, Dan, Srinivas, Tenenbaum, Kaelbling and Katz}]{PDDL_in_LLM}
\text{Silver, T.}, \text{Dan, S.}, \text{Srinivas, K.}, \text{Tenenbaum, J.~B.}, \text{Kaelbling, L.~P.} and \text{Katz, M.} (2023).
\newblock Generalized planning in pddl domains with pretrained large language models.
\newline\urlprefix\url{https://arxiv.org/abs/2305.11014}

\bibitem[{Song et~al.(2023)Song, Wu, Washington, Sadler, Chao and Su}]{LLMplanner}
\text{Song, C.~H.}, \text{Wu, J.}, \text{Washington, C.}, \text{Sadler, B.~M.}, \text{Chao, W.-L.} and \text{Su, Y.} (2023).
\newblock Llm-planner: Few-shot grounded planning for embodied agents with large language models.
\newline\urlprefix\url{https://arxiv.org/abs/2212.04088}

\bibitem[{Touvron et~al.(2023)Touvron, Lavril, Izacard, Martinet, Lachaux, Lacroix, Rozière, Goyal, Hambro, Azhar, Rodriguez, Joulin, Grave and Lample}]{Llama}
\text{Touvron, H.}, \text{Lavril, T.}, \text{Izacard, G.}, \text{Martinet, X.}, \text{Lachaux, M.-A.}, \text{Lacroix, T.}, \text{Rozière, B.}, \text{Goyal, N.}, \text{Hambro, E.}, \text{Azhar, F.}, \text{Rodriguez, A.}, \text{Joulin, A.}, \text{Grave, E.} and \text{Lample, G.} (2023).
\newblock Llama: Open and efficient foundation language models.
\newline\urlprefix\url{https://arxiv.org/abs/2302.13971}

\bibitem[{Vaswani et~al.(2023)Vaswani, Shazeer, Parmar, Uszkoreit, Jones, Gomez, Kaiser and Polosukhin}]{attention}
\text{Vaswani, A.}, \text{Shazeer, N.}, \text{Parmar, N.}, \text{Uszkoreit, J.}, \text{Jones, L.}, \text{Gomez, A.~N.}, \text{Kaiser, L.} and \text{Polosukhin, I.} (2023).
\newblock Attention is all you need.
\newline\urlprefix\url{https://arxiv.org/abs/1706.03762}

\bibitem[{Wei et~al.(2022)Wei, Tay, Bommasani, Raffel, Zoph, Borgeaud, Yogatama, Bosma, Zhou, Metzler, Chi, Hashimoto, Vinyals, Liang, Dean and Fedus}]{Emergent_Abilities}
\text{Wei, J.}, \text{Tay, Y.}, \text{Bommasani, R.}, \text{Raffel, C.}, \text{Zoph, B.}, \text{Borgeaud, S.}, \text{Yogatama, D.}, \text{Bosma, M.}, \text{Zhou, D.}, \text{Metzler, D.}, \text{Chi, E.~H.}, \text{Hashimoto, T.}, \text{Vinyals, O.}, \text{Liang, P.}, \text{Dean, J.} and \text{Fedus, W.} (2022).
\newblock Emergent abilities of large language models.
\newline\urlprefix\url{https://arxiv.org/abs/2206.07682}

\bibitem[{Wei et~al.(2023)Wei, Wang, Schuurmans, Bosma, Ichter, Xia, Chi, Le and Zhou}]{COT}
\text{Wei, J.}, \text{Wang, X.}, \text{Schuurmans, D.}, \text{Bosma, M.}, \text{Ichter, B.}, \text{Xia, F.}, \text{Chi, E.}, \text{Le, Q.} and \text{Zhou, D.} (2023).
\newblock Chain-of-thought prompting elicits reasoning in large language models.
\newline\urlprefix\url{https://arxiv.org/abs/2201.11903}

\bibitem[{Wolf et~al.(2020)Wolf, Debut, Sanh, Chaumond, Delangue, Moi, Cistac, Rault, Louf, Funtowicz, Davison, Shleifer, von Platen, Ma, Jernite, Plu, Xu, Scao, Gugger, Drame, Lhoest and Rush}]{huggingface}
\text{Wolf, T.}, \text{Debut, L.}, \text{Sanh, V.}, \text{Chaumond, J.}, \text{Delangue, C.}, \text{Moi, A.}, \text{Cistac, P.}, \text{Rault, T.}, \text{Louf, R.}, \text{Funtowicz, M.}, \text{Davison, J.}, \text{Shleifer, S.}, \text{von Platen, P.}, \text{Ma, C.}, \text{Jernite, Y.}, \text{Plu, J.}, \text{Xu, C.}, \text{Scao, T.~L.}, \text{Gugger, S.}, \text{Drame, M.}, \text{Lhoest, Q.} and \text{Rush, A.~M.} (2020).
\newblock Huggingface's transformers: State-of-the-art natural language processing.
\newline\urlprefix\url{https://arxiv.org/abs/1910.03771}

\bibitem[{Xie et~al.(2022)Xie, Raghunathan, Liang and Ma}]{ICL_explanation}
\text{Xie, S.~M.}, \text{Raghunathan, A.}, \text{Liang, P.} and \text{Ma, T.} (2022).
\newblock An explanation of in-context learning as implicit bayesian inference.
\newline\urlprefix\url{https://arxiv.org/abs/2111.02080}

\bibitem[{Yao et~al.(2023)Yao, Yu, Zhao, Shafran, Griffiths, Cao and Narasimhan}]{TOT}
\text{Yao, S.}, \text{Yu, D.}, \text{Zhao, J.}, \text{Shafran, I.}, \text{Griffiths, T.~L.}, \text{Cao, Y.} and \text{Narasimhan, K.} (2023).
\newblock Tree of thoughts: Deliberate problem solving with large language models.
\newline\urlprefix\url{https://arxiv.org/abs/2305.10601}

\bibitem[{Yao et~al.(2024)Yao, Li and Zhao}]{GOT}
\text{Yao, Y.}, \text{Li, Z.} and \text{Zhao, H.} (2024).
\newblock Beyond chain-of-thought, effective graph-of-thought reasoning in language models.
\newline\urlprefix\url{https://arxiv.org/abs/2305.16582}

\bibitem[{Zhang and Chai(2021)}]{HITUT}
\text{Zhang, Y.} and \text{Chai, J.} (2021).
\newblock Hierarchical task learning from language instructions with unified transformers and self-monitoring.
\newline\urlprefix\url{https://arxiv.org/abs/2106.03427}

\bibitem[{Zhang et~al.(2023)Zhang, Li, Cui, Cai, Liu, Fu, Huang, Zhao, Zhang, Chen, Wang, Luu, Bi, Shi and Shi}]{zhang2023sirenssongaiocean}
\text{Zhang, Y.}, \text{Li, Y.}, \text{Cui, L.}, \text{Cai, D.}, \text{Liu, L.}, \text{Fu, T.}, \text{Huang, X.}, \text{Zhao, E.}, \text{Zhang, Y.}, \text{Chen, Y.}, \text{Wang, L.}, \text{Luu, A.~T.}, \text{Bi, W.}, \text{Shi, F.} and \text{Shi, S.} (2023).
\newblock Siren's song in the ai ocean: A survey on hallucination in large language models.
\newline\urlprefix\url{https://arxiv.org/abs/2309.01219}

\bibitem[{Zhou et~al.(2019)Zhou, Pinto and Gupta}]{5655cf7f69e149148a2ea1c5a664b4ba}
\text{Zhou, W.}, \text{Pinto, L.} and \text{Gupta, A.} (2019).
\newblock Environment probing interaction policies.
\newblock Publisher Copyright: {\textcopyright} 7th International Conference on Learning Representations, ICLR 2019. All Rights Reserved.; 7th International Conference on Learning Representations, ICLR 2019 ; Conference date: 06-05-2019 Through 09-05-2019.

\end{thebibliography}
